\pdfoutput=1

\documentclass[10pt,journal,compsoc]{IEEEtran}
\ifCLASSOPTIONcompsoc
  \usepackage[nocompress]{cite}
\else
  \usepackage{cite}
\fi
\ifCLASSINFOpdf
  \usepackage[pdftex]{graphicx}
\else
\fi
\usepackage[cmex10]{amsmath}
\usepackage{amssymb}
\usepackage{array}

\ifCLASSOPTIONcompsoc
  \usepackage[caption=false,font=footnotesize,labelfont=sf,textfont=sf]{subfig}
\else
  \usepackage[caption=false,font=footnotesize]{subfig}
\fi
\usepackage{url}

\usepackage{paralist} 
\usepackage{cleveref} 

\newcommand{\eg}{e.\,g.~}
\newcommand{\ie}{i.\,e.~}
\newcommand{\etc}{etc~}
\renewcommand{\figurename}{Figure~}

\newcommand{\argmin}{\operatornamewithlimits{arg\ min}}

\begin{document}
%
\title{Predicting Face Recognition Performance Using Image Quality}
%
%
%
%

\author{Abhishek~Dutta,
        Raymond~Veldhuis,~\IEEEmembership{Senior~Member,~IEEE}
        and~Luuk~Spreeuwers,
\thanks{A. Dutta, R. Veldhuis and L. 
Spreeuwers are with the Department of EEMCS, University of Twente, Netherlands. E-mail: a.dutta@utwente.nl
\protect\\
see \texttt{http://www.abhishekdutta.org/phd-research/} for associated code and data.}
}

%
%

\ifCLASSOPTIONpeerreview
\markboth{Transactions on Pattern Analysis and Machine Intelligence,~Vol.~??, No.~?}%
{Dutta \MakeLowercase{\textit{et al.}}: Predicting Face Recognition Performance Using Image Quality}
\fi

%



\IEEEtitleabstractindextext{%
\begin{abstract}
This paper proposes a data driven model to predict the performance of a face recognition system based on image quality features.
We model the relationship between image quality features (\eg pose, illumination, \etc) and recognition performance measures using a probability density function.
To address the issue of limited nature of practical training data inherent in most data driven models, we have developed a Bayesian approach to model the distribution of recognition performance measures in small
regions of the quality space.
Since the model is based solely on image quality features, it can predict performance even before the actual recognition has taken place.
We evaluate the performance predictive capabilities of the proposed model for six face recognition systems (two commercial and four open source) operating on three independent data sets: MultiPIE, FRGC and CAS-PEAL.
Our results show that the proposed model can accurately predict performance using an accurate and unbiased Image Quality Assessor (IQA).
Furthermore, our experiments highlight the impact of the unaccounted quality space -- the image quality features not considered by IQA -- in contributing to performance prediction errors.
\end{abstract}

\begin{IEEEkeywords}
Image Quality, Performance Prediction, Predictive Models, Image Quality Features
\end{IEEEkeywords}}

\maketitle

\IEEEdisplaynontitleabstractindextext

%
\IEEEpeerreviewmaketitle

\IEEEraisesectionheading{\section{Introduction}\label{sec:introduction}}

%
%
%
%
\IEEEPARstart{A}{face} verification system compares a pair of facial images and decides whether the image pair is a match (originating from the same individual) or non-match (originating from different individuals) based on their similarity score which is compared with a verification decision threshold.
Given that practical face recognition systems make occasional mistakes in such verification decisions, there is a need to quantify the uncertainty of decision about identity.
In other words, we are not only interested in the verification decision (match or non-match) but also in its uncertainty.

\begin{figure}
 \centering
 \includegraphics[width=0.8\linewidth]{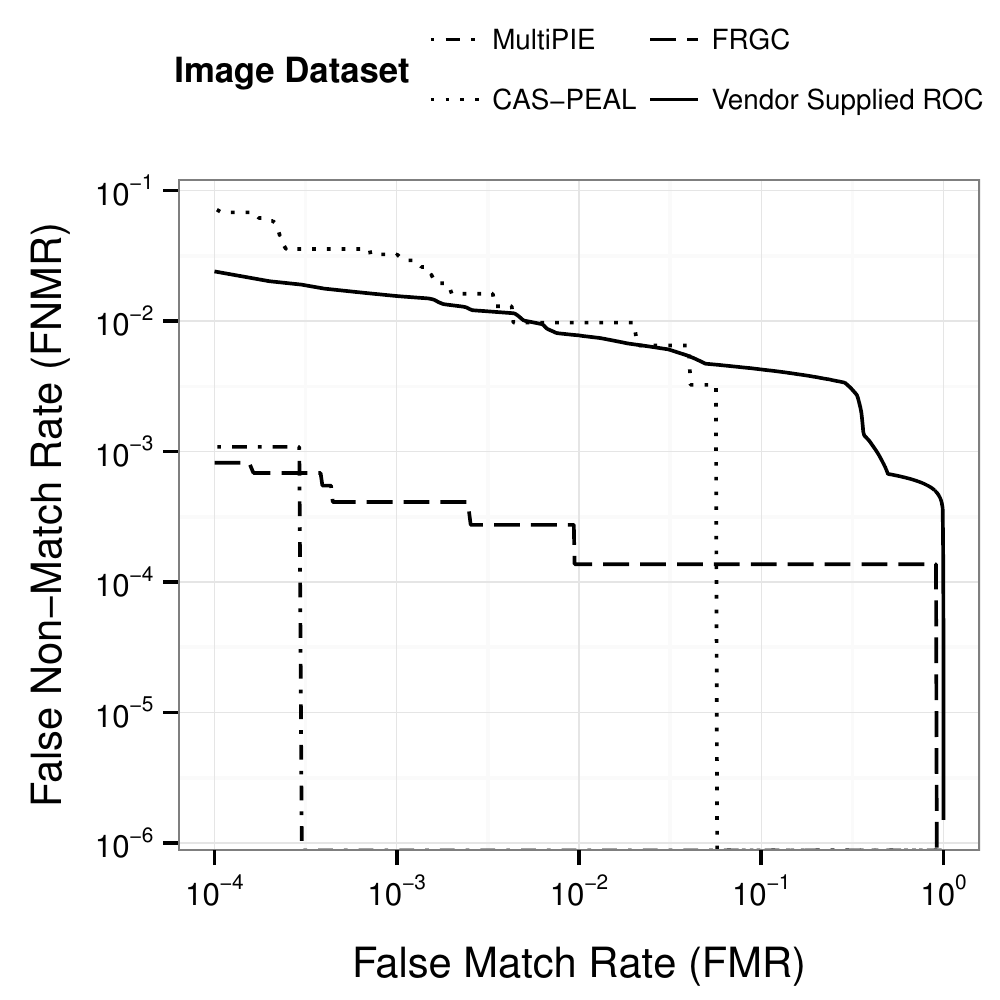}
 \caption{Vendor supplied Receiver Operating Characteristic (ROC) and actual ROC curve of a COTS face recognition system~\cite{facevacs2010} operating on frontal pose, illumination, neutral expression subset of three independent data sets (sample facial images are shown in~\figurename~\ref{fig:prb_ref_img_illus.pdf}).}
 \label{fig:fv_vendor_roc_vs_true_roc.pdf}
\end{figure}

The vendors of commercial off-the-shelf (COTS) face recognition systems provide the Receiver Operating Characteristics (ROC) curve which characterizes the uncertainty of the decision about identity at several operating points in terms of trade-off between false match and false non-match rates.
As shown in~\figurename\ref{fig:fv_vendor_roc_vs_true_roc.pdf}, the vendor supplied ROC for a COTS face recognition system~\cite{facevacs2010} differs significantly from ROCs obtained from frontal image subsets of three facial image data sets~\cite{gross2008multipie,phillips2005overview,gao2008caspeal} that were captured using different devices and under different setup.
Usually, the vendor supplied ROC represents recognition performance that the face recognition system is expected to deliver under ideal conditions.
In practice, the ideal conditions are rarely met and therefore the actual recognition performance varies as illustrated in~\figurename\ref{fig:fv_vendor_roc_vs_true_roc.pdf}.
Therefore, practical applications of verification systems cannot rely on the vendor supplied ROC curve to quantify uncertainty in decision about identity on per verification instance basis.

In this paper, we address this problem by presenting a generative model that predicts the verification performance based on image quality.
In addition to the inherent limitations of a face recognition system, the quality (like pose, illumination direction, noise, etc) of the pair of facial images used in verification process also contribute to the uncertainty in decision about identity.
For example, a verification decision made using a non-frontal image with uneven lighting entails more uncertainty than a verification decision carried out on frontal mugshots captured under studio conditions.
Therefore, in this paper, we use image quality as the feature for predicting performance of a face recognition system.
Throughout this paper, we use the term ``image quality'' to refer to all the static or dynamic characteristics of the subject or acquisition process as described in~\cite{iso_iec_29794-5:2010}, including for instance facial pose, illumination direction, \etc.

A large and growing body of literature has investigated the use of similarity scores (\ie classifier's output) as a feature for performance prediction.
However, there is evidence that non-match scores are influenced by both facial identity and the quality of image pair~\cite{dutta2013facial}.
Therefore, it is not possible to discern if a low non-match score is due to non-match identity or poor quality of image pair.
Hence, we avoid using similarity scores as a performance feature in the proposed solution.
This design decision not only avoids the issues associated with using similarity score as a feature but also allows our model to predict performance even before the actual facial comparison has taken place.

A substantial amount of literature has tried to model the similarity score distribution (match and non-match) conditioned upon image quality in order to predict performance\cite{shi2008modeling,wein2005using,scheirer2011meta}.
Since the parameter of interest (\ie similarity score) is a uni-variate variable, the model is much simpler and any recognition performance measure can be derived from these models of score distributions.
In practice, we rarely need to know about the underlying score distributions and are mostly interested in the recognition performance that can be expected from a particular face recognition system operating on a given image pair.
Therefore, in this paper, we take a more practical approach of directly modeling the recognition performance measure (\eg False Non-Match Rate - FNMR and False Match Rate - FMR at a certain point of operation) of interest rather than modeling intermediate variable (\ie similarity score).
The proposed model is flexible to accommodate any type of recognition performance measure that is of interest to the user like Area Under ROC (AUC), Equal Error Rate (EER), calibrated log-likelihood-ratios, \etc.

There are many applications of models that can predict the performance of a face recognition system.
In forensic cases involving face recognition, it can rank CCTV footage frames based on the image quality of each frame thereby allowing forensic investigators to focus their effort on a smaller set of images with higher evidential value.
When capturing facial images for the reference set (\ie enrollment or gallery set), it can alert the operator whenever a ``poor'' quality image sneaks into the enrollment set.
Such a model can be used to dynamically set the verification decision threshold that adapts according to the sample quality, for instance to maintain a prescribed False Match Rate (FMR).
The tolerance of face recognition algorithms to image quality degradation varies and therefore results from multiple algorithms can be fused based on the predicted performance corresponding to individual face recognition algorithm.

The method we present is based on modeling the relationship between image quality and face recognition performance using a probability density function.
During the training phase, this density function is approximated by evaluating the recognition performance corresponding to the quality variations encountered in practice -- a data driven approach.
A model of this density function learned during the training phase allows us to predict the performance of a face recognition system on previously unseen facial images even before the actual verification has taken place.

This paper is organized as follows:
We review some of the existing literature on performance prediction in Section~\ref{dutta2015predicting_related_work}.
Section~\ref{dutta2015predicting_model_description} describes the proposed generative model which uses image quality features to predict performance of a face recognition system.
In Section~\ref{dutta2015predicting_exp}, we present the result of model training and performance prediction on three independent data sets for six face recognition systems.
The key observations from these experiments are discussed in Section~\ref{dutta2015predicting_discussion} followed by final conclusions in Section~\ref{dutta2015predicting_conclusion}.

\section{Related Work}
\label{dutta2015predicting_related_work}

\begin{figure*}
\centering
\includegraphics[width=\linewidth]{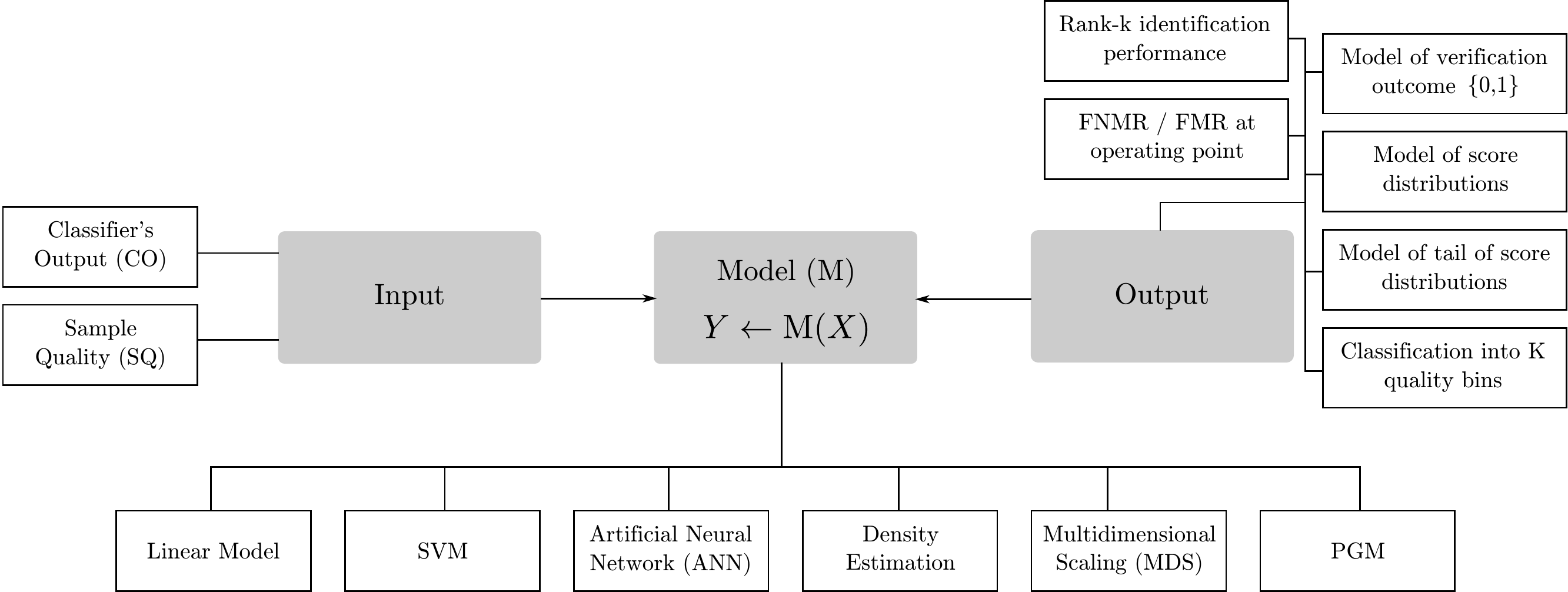}
\caption{Typical components of a system aiming to predict the performance of a biometric system.}
\label{fig:biometric_perf_pred_literature_map.pdf}
\end{figure*}

Systems aiming to predict the recognition performance are characterised by three components: \textit{Input} denotes the features with performance predictive capability; \textit{Output} denotes the  recognition performance measure of interest; and \textit{Model} corresponds to a model that represents the functional relationship between Input and Output.
The existing publications on performance prediction differ in the variants of Input, Model and Output as listed in~\tablename~\ref{tbl:literature_classification}. In~\figurename\ref{fig:biometric_perf_pred_literature_map.pdf}, we show all the variants of these components that we found during literature review.

In this paper, we classify the existing literature into two groups based on the type of feature (\ie Input) used for performance prediction.
The first group of performance prediction systems use output of the classifier (CO) itself as a feature for performance prediction while the second group uses biometric sample quality (SQ).

\begin{table}[h]
\renewcommand{\arraystretch}{1.3}
\caption{Classification of existing literature on performance prediction based on the three typical components of such systems as shown~\figurename\ref{fig:biometric_perf_pred_literature_map.pdf}.}
\label{tbl:literature_classification}
\centering
\footnotesize
\begin{tabular}{p{1.2cm} | p{1.0cm} | p{1.8cm} | p{2.9cm}}
\hline
Paper & Input & Output & Model \\
\hline
\cite{li2005predicting} & CO & Rank-k & SVM \\
\cite{wang2007modeling} & CO & Rank-k & SVM \\
\cite{scheirer2011meta} & CO & Model of tail & Density Est. (Wbl) \\
\cite{shi2008modeling} &  CO & Model of tail & Density Est. (GPD) \\
\cite{klare2012face} & CO & Score dist. & Density Est. (KDE) \\
\cite{ozay2009improving} & CO & Verification & PGM \\
\hline
\cite{beveridge2008focus},~\cite{beveridge2010quantifying} & SQ & FNMR/FMR & Linear Model (GLMM) \\
\cite{dutta2014bayesian} & SQ & FNMR/FMR & Density Est. (GMM) \\
\textbf{our work} & \textbf{SQ} & \textbf{FNMR/FMR} & \textbf{Density Est. (GMM)} \\
\cite{wein2005using} & SQ & Score dist. & Density Est. ($\gamma$, $\textnormal{ln} \; \mathcal{N}$) \\
\cite{aggarwal2011predicting} & SQ & Score dist. & MDS \\
\cite{aggarwal2012predicting} & SQ & Quality bins & Linear Model (PLS) \\
\cite{tabassi2004fingerprint},~\cite{tabassi2005novel} & SQ & Quality bins & ANN \\
\hline
\cite{jammalamadaka2012algorithm}* & CO/SQ & pred. pose err. & SVM \\
\cite{zuo2010adaptive}* & CO/SQ & Verification & ANN \\

\hline
\end{tabular}
\vspace{0.2cm}

* denotes work in domains other than face and fingerprint biometrics \\
\end{table}

The key observation underpinning the first group of existing literature is that the overlapping region between match and non-match score distribution entail more uncertainty in decision about identity.
They begin by creating features from classifier's output (\ie similarity score) that are predictive of recognition performance.
For example,~\cite{li2005predicting} create a set of three features based on similarity score while~\cite{wang2007modeling} uses similarity score based features to quantify the intrinsic factors (properties of algorithm, reference set, \etc) and extrinsic factors (properties of probe set).
Rather than considering the full set of similarity scores, \cite{scheirer2011meta}~and~\cite{shi2008modeling} argue that decision about identity is more uncertain in the overlapping region of the match and non-match distributions and therefore they consider the similarity scores only in the tail region.
The authors of~\cite{ozay2009improving} use the distance of a similarity score from the non-match distribution in units of standard deviation (\ie d-prime value~\cite{jain2007handbook}) while \cite{klare2012face} use the facial uniqueness feature derived from the nature of subject specific non-match distribution as the performance predictor feature.
A major limitation of using features derived from similarity scores is that they become unstable under quality variations~\cite{dutta2013facial}.

The second group of existing literature is based on the observation that sample quality influences the uncertainty in decision about identity -- empirical evidence show that poorer sample quality entails more uncertainty in decision about identity.
They begin by externally assessing image quality features of facial images using an Image Quality Assessor (IQA).
For instance,~\cite{tabassi2004fingerprint} and \cite{tabassi2005novel} use fingerprint image quality like clarity of ridges and valleys, number and quality of minutiae, size of image, \etc while~\cite{wein2005using} use fingerprint quality assessments from a propriety IQA as image quality features.
The authors of~\cite{aggarwal2012predicting} use image-specific (like image sharpness, image hue content, image saturation, \etc) and face-specific (like expression) characteristics as image quality features.
A single image quality feature that characterizes the nature of illumination in a facial image was used in~\cite{aggarwal2012predicting}.
Using the term co-variate to denote image quality, \cite{beveridge2008focus}~and~\cite{beveridge2010quantifying} use a wide range of subject co-variates like age, gender, race, wearing glasses and image co-variates like focus, resolution, head tilt as the features for performance prediction.
A major limitation of using image quality as a performance prediction feature is that there are overwhelmingly large number of quality factors that may influence the performance of a face recognition system -- their exact count is still unknown.
Furthermore, accurate measurement of image quality is still an unsolved problem and concerted efforts (like NFIQ2~\cite{nfiq2}) are underway to develop an extensive set of quality feature and to standardize the use and exchange of quality measurements.
The authors of~\cite{phillips2013existence} have proposed the Greedy Pruned Ordering (GPO) scheme to determine the best case upper bound performance prediction capability that can be achieved by any quality measure on a \textit{particular combination of algorithm and data set}.

Some existing works like~\cite{jammalamadaka2012algorithm} and \cite{zuo2010adaptive} belong to both the first and second group because they combine both classifier's output (CO) and image quality features (SQ) to predict performance.

The choice of recognition performance measure (\ie Output) is based on user requirements.
For instance, Rank-k recognition rate and FNMR/FMR at the operating point are the recognition performance measure used for modeling identification and verification performance respectively.
Authors choosing to model the similarity score distribution do not need to define the recognition performance measure because any performance measure can be derived from the model of similarity score distribution.
Some authors model discrete quality bins with distinctly different recognition performance as the output.
In this paper, we model the following recognition performance measure: FNMR and FMR at a particular operating point defined by a decision threshold.
Some existing works like~\cite{ozay2009improving} and \cite{zuo2010adaptive} have tried to directly predict the success/failure of the verification outcome which according to~\cite{phillips2009introduction} is a pursuit equivalent to finding a perfect verification system.

Once the performance predictor feature (\ie Input) and the desired recognition performance measure (\ie Output) is fixed, the final step is to use an appropriate model to learn the relationship between predictor features and recognition performance.
So far, many variants of learning algorithms has been applied to learn the relationship between performance predictor features and the recognition performance measure.
For instance, \cite{wang2007modeling} and \cite{li2005predicting} use Support Vector Machine (SVM) to model this relationship while \cite{tabassi2004fingerprint} and \cite{tabassi2005novel} use the Artificial Neural Network (ANN) to learn the relationship between fingerprint sample quality features and the normalized similarity score -- the distance of match score from non-match score distribution.
The authors aiming to model similarity score distributions conditioned on image quality either use a standard parametric distribution like Weibull~\cite{scheirer2011meta}, General Pareto Distribution (GPD)~\cite{shi2008modeling}, gamma/log-normal distributions~\cite{wein2005using} or use Kernel Density Estimation (KDE)~\cite{klare2012face} when the score distribution cannot be explained by standard parametric distributions.
The authors of~\cite{aggarwal2011predicting} apply Multi-Dimensional Scaling (MDS) to model the relationship between quality features and match score distribution while in~\cite{aggarwal2012predicting}, the authors use regression to model the relationship between quality partition (good, bad and ugly) and image quality features.

In this paper, we extend the work of~\cite{dutta2014bayesian} in many fronts.
We address the issue of limited training data set by using a probabilistic model of quality and recognition performance in small regions of the quality space.
We also report the accuracy of predicted performance on three independent facial image data sets for six face recognition systems.
We use the conditional expectation, instead of maximum a posteriori probability (MAP), to estimate the recognition performance given the image quality.
We also present performance prediction results using an unbiased IQA.
Furthermore, our work most closely relates to the work of~\cite{beveridge2008focus} which uses a Generalized Linear Mixed Model (GLMM) to model the relationship between image quality (like focus, head tilt, \etc) and the False Non-Match Rate (FNMR) at a given False Match Rate (FMR).
Their analysis focused on investigating the impact of each quality metric on recognition performance.
Our work focuses on predicting performance for a given sample quality.

\section{Model of Image Quality and Recognition Performance}
\label{dutta2015predicting_model_description}
Let $\mathbf{q}^{p} \in \mathbb{R}^{m}$ and $\mathbf{q}^{g} \in \mathbb{R}^{m}$ be the vectors denoting the image quality features (like pose, illumination direction, noise, \etc) of a probe and reference image pair respectively as assessed by an Image Quality Assessment (IQA) system.
We coalesce $\mathbf{q}^{p}$ and $\mathbf{q}^{g}$ to form a single quality feature vector $\mathbf{q}=[\mathbf{q}^{p}; \mathbf{q}^{g}] \in \mathbb{R}^{2m}$ which denotes the image quality features of probe and reference image pair.
For a particular face recognition system, let $\mathbf{r} \in \mathbb{R}^{n}$ denote the face recognition performance corresponding to a sufficiently large set of different probe (or, query) and reference (or, enrollment) image pairs having same image quality $\mathbf{q}$.
The prosed model is flexible to accommodate any recognition performance parameter of interest to the user in vector $\mathbf{r}$.
For instance, if we wish to model and predict the Receiver Operating Characteristics (ROC) curve of a face recognition system, then vector $\mathbf{r}$ would correspond to FMR and FNMR measure sampled at several operating points.
Other recognition performance measures like Area Under Curve (AUC), Equal Error Rate (EER), calibrated log-likelihood-ratios, \etc can fit equally well in vector $\mathbf{r}$.
Here, we assume that vector $\mathbf{q}$ is sufficient to capture all the relevant quality variations possible in a facial image pair that have an influence on face recognition performance.
Different face recognition systems have varying level of tolerance to image quality degradations and therefore vector $\mathbf{r}$ is a function of a particular face recognition system.

\begin{figure}
 \centering
 \includegraphics[width=0.8\linewidth]{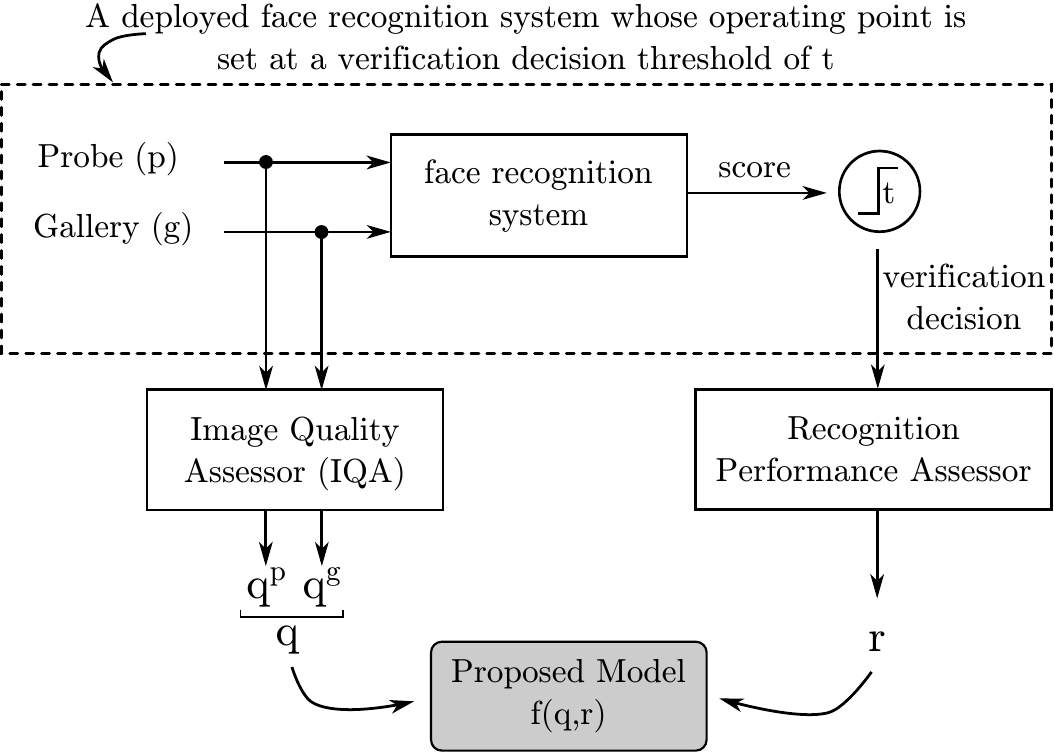}
 \caption{The proposed performance prediction model treats a face recognition system as a ``black box'' and captures the relationship between image quality features $\mathbf{q}$ and recognition performance measures $\mathbf{r}$ using a probability density function $f(\mathbf{q},\mathbf{r})$.}
 \label{fig:black_box_qr_model_illus}
\end{figure}

To model the interaction between image quality features $\mathbf{q}$ and recognition performance $\mathbf{r}$, we coalesce $\mathbf{q}$ and $\mathbf{r}$ and form the Quality-Performance (QR) space.
We model this QR space using a Probability Density Function (PDF) $f(\mathbf{q},\mathbf{r})$ as depicted in~\figurename\ref{fig:black_box_qr_model_illus}.
This PDF defines the probability of observing certain combination of image quality $\mathbf{q}$ and recognition performance $\mathbf{r}$.
Given the quality $\mathbf{q}$ of previously unseen verification instance, we can apply the Bayes' theorem to obtain the posterior distribution of recognition performance $\mathbf{r}$ as follows:
\begin{equation}
f(\mathbf{r}|\mathbf{q}) = \frac{f(\mathbf{q},\mathbf{r})}{f(\mathbf{q})}.
\label{eq:P_r_given_q}
\end{equation}
The conditional expectation of $\mathbf{r}$ with respect to the conditional probability distribution of~\eqref{eq:P_r_given_q} is:
\begin{equation}
\mathbb{E}(\mathbf{r}|\mathbf{q}) = \int \mathbf{r} f(\mathbf{r}|\mathbf{q}) dr,
\label{eq:expectation_r}
\end{equation}
where, $\mathbb{E}(\mathbf{r}|\mathbf{q})$ denotes the expected value of recognition performance for a given image quality pair $\mathbf{q}$.
In this paper, we use $\mathbb{E}(\mathbf{r}|\mathbf{q})$ as an estimate of recognition performance $\mathbf{r}$ given quality features $\mathbf{q}$ of probe and reference image pair.

\subsection{Model Training : Estimating $f(\mathbf{q},\mathbf{r})$ from data}
\label{model_training}
In this paper, we model the Probability Density Function (PDF) $f(\mathbf{q},\mathbf{r})$ using a mixture of $K$ multivariate Gaussian (MOG)
\begin{equation}
f(\mathbf{q},\mathbf{r}) = \sum_{k=1}^{K} \pi_k f_{k}(\mathbf{q},\mathbf{r}),
\label{eq:P_q_r_MOG}
\end{equation}
where, $f_{k}(\mathbf{q},\mathbf{r})=\mathcal{N}([\mathbf{q},\mathbf{r}]; \mu_k, \Sigma_k)$ denotes the $k^{\textnormal{th}}$ Gaussian mixture component with mean $\mu_k$ and covariance $\Sigma_k$ and $\pi_k$ are the mixture coefficients such that $0~\leq~\pi_k~\leq~1$, $\sum_{k}~\pi_k~=~1$.

To learn the parameters of MOG in~\eqref{eq:P_q_r_MOG}, we require a training data set $\mathcal{D}_{\textnormal{train}}=\{ [\mathbf{q}_{i}, \mathbf{r}_{i}] \}$ where each $\mathbf{q}_{i}$ denotes a sample point in quality space and $\mathbf{r}_{i}$ is the corresponding recognition performance.
Such a training data set can be created only if we have sufficiently large number of similarity scores (both match and non-match) at each sampling point $\mathbf{q}$ in the quality space.
In other words, for each point of interest in quality space $\mathbf{q}$, we require a training data set with a large number of unique verification attempts having probe and reference image quality $\mathbf{q}$.
In practice, it is very difficult to obtain such a training data set.
Due to limited nature of practical training data, we cannot reliably evaluate recognition performance at discrete points in the quality space.
Therefore, we build probabilistic models of quality and performance in small regions of the quality space as described in~Section~\ref{prob_model_q_r}.
We randomly sample from these models of $\mathbf{q}$ and $\mathbf{r}$ to build the training data set $\tilde{\mathcal{D}}_{\textnormal{train}} = \{ [\tilde{\mathbf{q}_{i}}, \tilde{\mathbf{r}_{i}}] \}$.
This strategy of building the training data set allows us to capture the uncertainty in quality and performance measurements entailed by IQA and limited training data set respectively.

The size and location of small regions in the quality space is determined by the nature of training data which commonly has densely populated samples in the regions of quality space corresponding to most common types of quality variations and sparse samples in other regions.
We define $N_{\textnormal{qs}}$ quantile points (along each quality axis) based on quantiles of evenly space probabilities.
Unique sampling points are formed in the quality space by taking all the possible combination of these $N_{\textnormal{qs}}$ quantile points along each quality axis.
We form regions around these quality space sampling points such that the adjacent quantile points define the boundary of these overlapping regions.


The raw training data is composed of unique verification attempts and therefore each record of the training data set contains: similarity score $s$, quality of probe and reference images $\mathbf{q}$ and the ground truth (match or non-match) of the verification decision.
Let $\mathbf{Q}(\mathbf{q})$ and $\mathbf{S}(\mathbf{q})$ denote the set of quality samples and corresponding similarity scores respectively present in a quality space region formed around a quality space sampling point.
As described in Section~\ref{prob_model_q_r}, we build a probabilistic model of $\mathbf{q}$ and $\mathbf{r}$ in each quality space region and randomly sample $N_{\textnormal{rand}}$ samples from these models to create the data set $\tilde{\mathcal{D}}_{\textnormal{train}} = \{ [\tilde{\mathbf{q}}_{i}, \tilde{\mathbf{r}}_{i}] \}$ needed to train the model of (\ref{eq:P_q_r_MOG}).
We pool the QR training data from each quality space region and apply the Expectation Maximization (EM) algorithm~\cite{fraley2012mclust} to learn the parameters $(\pi_{k}, \mu_{k}, \Sigma_{k})$ of Mixture of Gaussian in~\eqref{eq:P_q_r_MOG}.

\subsubsection{Probabilistic Model of quality $\mathbf{q}$ and performance $\mathbf{r}$}
\label{prob_model_q_r}
We now describe the probabilistic model of $\mathbf{q}$ and $\mathbf{r}$ in the quality space region containing quality samples $\mathbf{Q}(\mathbf{q})$ and similarity score samples $\mathbf{S}(\mathbf{q})$.
The random samples from these probabilistic models form the training data set used to learn the model parameters of~\eqref{eq:P_q_r_MOG}.

We assume that the elements of $\mathbf{q}$ are mutually independent and they follow a Gaussian distribution within the quality region.
Based on this assumption, we fit a multivariate Gaussian $\mathcal{N}(\mathbf{q}; \mu_{i}, \Sigma_{i}^{\textnormal{diag}})$ with diagonal covariance matrix parametrization to all quality samples in $\mathbf{Q}$.

Now, we describe a probabilistic model for recognition performance in the quality region.
First, we define the recognition performance measure used throughout this paper.
For face recognition systems deployed in real-world, the operating point is set to achieve certain False Match Rate (FMR) or False Non-Match Rate (FNMR).
In this paper, we assume that the recognition performance $\mathbf{r}$ of interest is the FMR and FNMR at certain decision threshold $t$ (which defines the operating point of a face recognition system): $\mathbf{r} = [\textnormal{FMR}_{t}, \textnormal{FNMR}_{t}]$.

Given an observation of the number of similarity scores below (and above) the decision threshold $t$, we want to know the nature of the distribution of FNMR (and FMR).
First, we consider the set of match scores $\mathbf{M}(\mathbf{q}) \subset \mathbf{S}(\mathbf{q})$ to build a model of FNMR distribution.
Each element in $\mathbf{M}(\mathbf{q})$ is a similarity score corresponding to a pair of probe and reference image containing facial images of same subject (\ie match pair).
For all the elements in $\mathbf{M}(\mathbf{q})$, we can make a verification decision $w \in \{0,1\}$ based on the decision threshold $t$ as follows
\begin{equation}
w_{j} = 
\begin{cases}
1 & \text{if} \; M(\mathbf{q})[j] < t, \\
0 & \text{otherwise},
\end{cases}
\label{eq:bernouli_trial_expression}
\end{equation}
where, $M(\mathbf{q})[j]$ corresponds to the $j^{\textnormal{th}}$ similarity score in set $\mathbf{M}(\mathbf{q})$ and $w_{j}=1$ is used to denote failure in verification of a match pair (\ie False Non-Match) while $w_{j}=0$ denotes success in verification of a math pair.
Therefore, each verification decision $w_{j}$ can be thought of as the outcome of a Bernoulli trial where success and failure corresponds to $w_{j}=1$ and $w_{j}=0$ respectively.
Let $\mathbf{w}=\{w_{j}\}$ denote a Binomial experiment containing a set of $N=|\mathbf{M}(\mathbf{q})|$ statistically independent Bernoulli trials such that $f(w_{j}=1|N, \tau)= \tau$ where $\tau$ is the probability of failure in verification of a match pair which is also called the False Non-Match Rate (FNMR).
Furthermore, let $m$ be a random variable indicating the number of $w_{j}=1$ (\ie success) in the Binomial experiment.
The value of False Non-Match Rate (FNMR) is given by:
\begin{equation}
\textnormal{FNMR} = \frac{m}{N}.
\end{equation}
We are interested in the distribution of FNMR which in turn depends on the distribution of random variable $m$.
The probability of getting $m$ success in $N$ trials follows a Binomial distribution defined as follows
\begin{equation}
\textnormal{Bin}(m|N,\tau) = \binom{N}{m} \tau^{m} (1-\tau)^{N-m},
\label{eq:binomial_likelihood}
\end{equation}
where, $\tau$ denotes the probability of getting success in a Bernoulli trial (\ie FNMR).
Taking a Bayesian perspective on the problem of estimating distribution of $\tau$, we first define the prior distribution over $\tau$.
Since, (\ref{eq:binomial_likelihood}) belongs to the exponential family, we chose the Beta distribution $\textnormal{Beta}(\tau|a,b)$ as the prior for $\tau$ where $a,b$ denote the shape parameters of the Beta distribution.
Based on the property of conjugate priors~\cite[p.70]{bishop2006pattern}, the posterior distribution, which is also a Beta distribution, is
\begin{equation}
f(\tau|m,l,a,b) = \textnormal{Beta}(m+a, l+b) \\
\label{eq:tau_density}
\end{equation}
where, $l=N-m$ denotes the number failures in $N$ Bernoulli trial.
This shows that the underlying uncertainty in FNMR is given by a Beta distribution.
In a similar way, we can show that the uncertainty in FMR is also given by a Beta distribution.

In order to create the training data set $\tilde{\mathcal{D}}_{\textnormal{train}} = \{ [\tilde{\mathbf{q}}_{i}, \tilde{\mathbf{r}}_{i}] \}$, we draw $N_{\textnormal{rand}}$ random samples independently from the multivariate Gaussian distribution model of $\mathbf{q}$ and Beta distributions corresponding to FMR and FNMR.

Since we do not have any prior knowledge about the distribution of FMR and FNMR in a quality region, we assume a uniform prior \ie $\textnormal{Beta}(1,1)$.
Furthermore, since FMR and FNMR values follow a Beta distribution, the recognition performance measure $r_{i}$ has a Bayesian credible interval $(c, d)$ of size $1-\alpha$ such that
\begin{equation}
\int_{c}^{d} \textnormal{Beta}(r; a,b) \; dr = 1 - \alpha
\label{eq:credible_interval}
\end{equation}

\subsection{Performance Prediction}
Given a previously unseen probe and reference image pair with quality $\mathbf{q}$, we now derive an expression for the posterior distribution of recognition performance $f(\mathbf{r}|\mathbf{q})$.

From training, we have a model $f(\mathbf{q},\mathbf{r})$ defined as 
\begin{equation}
\begin{aligned}
f(q,r) &= \sum_{k} \pi_k \mathcal{N}([\mathbf{q},\mathbf{r}]; \mu_k, \Sigma_k) \\
 &= \sum_{k} \pi_k f_{k}(\mathbf{q},\mathbf{r}).
\end{aligned}
\label{eq:f_q_r}
\end{equation}

The marginal distribution $f(\mathbf{q})$ is given by
\begin{align}
f(\mathbf{q}) &= \int_{r} f(\mathbf{q},\mathbf{r}) \; dr \nonumber \\
 &= \int_{r} \sum_{k} \pi_k f_{k}(\mathbf{q},\mathbf{r}) \; dr \qquad \textnormal{from~\eqref{eq:f_q_r}} \nonumber \\
 &= \sum_{k} \pi_k \int_{r} f_{k}(\mathbf{q},\mathbf{r}) \; dr \qquad \textnormal{since $\pi_k f_{k}(\mathbf{q},\mathbf{r}) \geq 0$} \nonumber \\
 &= \sum_{k} \pi_k f_{k}(\mathbf{q})
\label{eq:f_q}
\end{align}

For a given quality $\mathbf{q}$, the conditional distribution of $\mathbf{r}$ is obtained by applying the Bayes' theorem as follows
\begin{align}
f(\mathbf{r}|\mathbf{q}) = \frac{f(\mathbf{r},\mathbf{q})}{f(\mathbf{q})}.
\label{eq:f_r_given_q1}
\end{align}
Substituting~\eqref{eq:f_q_r}~and~\eqref{eq:f_q}~in~\eqref{eq:f_r_given_q1},
\begin{align}
f(\mathbf{r}|\mathbf{q}) &= \frac{\sum_{} \pi_k f_{k}(\mathbf{q},\mathbf{r})}{\sum_{} \pi_k f_{k}(\mathbf{q})}.
\label{eq:f_r_given_q2}
\end{align}
Applying the Bayes's theorem to $f_{k}(\mathbf{q},\mathbf{r})=f_{k}(\mathbf{r}|\mathbf{q}) f_{k}(\mathbf{q})$ in~\eqref{eq:f_r_given_q2}, the posterior distribution of $\mathbf{r}$ for the given quality $\mathbf{q}$ is
\begin{align}
f(\mathbf{r}|\mathbf{q}) &= \frac{\sum_{k} \pi_k f_{k}(\mathbf{r}|\mathbf{q}) f_{k}(\mathbf{q})}{\sum_{k} \pi_k f_{k}(\mathbf{q})} \nonumber \\
 &= \sum_{k} f_{k}(\mathbf{r}|\mathbf{q}) \left( \frac{\pi_k f_{k}(\mathbf{q})}{\sum_{k} \pi_k f_{k}(\mathbf{q})} \right) \nonumber \\
 &= \sum_{k} \psi_{k} f_{k}(\mathbf{r}|\mathbf{q})
\label{eq:f_r_given_q3}
\end{align}

where, $\psi_{k}$ denotes the new weights for conditional mixture of Gaussian.
The conditional and marginal distribution of each mixture component is given by \cite{petersen2008matrix}:
\begin{equation*}
\begin{aligned}
f_{k}(\mathbf{r}|\mathbf{q}) &= \mathcal{N}([\mathbf{q},\mathbf{r}]; \hat{\mu}_{k,r}, \hat{\Sigma}_{k,r}) \\
f_{k}(\mathbf{q}) &= \mathcal{N}([\mathbf{q}]; \mu_{k,q}, \Sigma_{k,q})
\end{aligned}
\end{equation*}
where, 
\begin{equation*}
\begin{aligned}
\hat{\mu}_{k,r} &= \mu_{k,r} + \Sigma_{k,c}^{T} \Sigma_{k,q}^{-1} (\mathbf{q} - \mu_{k,q}) \\
\hat{\Sigma}_{k,r} &= \Sigma_{k,r} - \Sigma_{k,c}^{T} \Sigma_{k,q}^{-1} \Sigma_{k,c} \\
\mu_{k} &= \begin{bmatrix} \mu_{k,q} \\ \mu_{k,r} \end{bmatrix} \\
\Sigma_{k} &= \begin{bmatrix} \Sigma_{k,q} & \Sigma_{k,c} \\ \Sigma_{k,c}^{T} & \Sigma_{k,r} \end{bmatrix}
\end{aligned}
\end{equation*}

Using~\eqref{eq:expectation_r}, the estimate of recognition performance $\mathbf{r}$ is given by the conditional expectation as follows:
\begin{align}
\mathbb{E}(\mathbf{r}|\mathbf{q}) &= \sum_{k} \psi_{k} \mathbb{E} ( f_{k}(\mathbf{r}|\mathbf{q}) ) \qquad \textnormal{from~(\ref{eq:f_r_given_q3})} \nonumber \\
 &= \sum_k \psi_{k} \; \hat{\mu}_{k,r}
\label{eq:expectation_r_given_q0}
\end{align}

In words, estimate of the predicted recognition performance is equal to the weighed sum of conditional mean of each conditional mixture component.

\subsection{Model Parameter Selection}
\label{model_param_selection}
The proposed Gaussian Mixture model of~(\ref{eq:P_q_r_MOG}), requires selection of the following two parameters:
\begin{inparaenum}[\itshape a\upshape)]
\item Number of mixture components $K$,
\item Parametrization $\Sigma^{p}$ of the covariance matrix $\Sigma$.
\end{inparaenum}
We require parametrization of the covariance matrix because we lack sufficient training data to estimate the full covariance matrix.
Let $\theta = [K, \Sigma^{p}]$ denote the parameter for our optimization, where
\begin{align}
\Sigma^{p} & \in \{ \text{EII}, \text{VII}, \text{EEI}, \text{VEI}, \text{EVI}, \text{VVI}, \text{EEE}, \text{EEV}, \text{VEV}, \text{VVV} \} \nonumber \\
K & \in [ \textnormal{k}_{min}, \textnormal{k}_{max} ] \nonumber
\end{align}
and, the search space for $\Sigma^{p}$ is based on the covariance matrix parametrization scheme presented in~\cite{fraley2012mclust}.
In this parametrization scheme, the three characters denote volume, shape and orientation respectively of the mixture components.
Furthermore, E denotes equal, V means varying across mixture components and I refers to identity matrix in specifying shape or orientation.
For example, EVI corresponds to a model in which all mixture components have equal (E) volume, the shapes of mixture components may vary (V) and the orientation is the identity (I).

We select the optimal model parameter based on the Bayesian Information Criterion (BIC)~\cite{fraley2012mclust} defined as:
\begin{equation}
\textnormal{BIC} \equiv 2 \; \textnormal{ln} \; f(q,r|\theta) - n \; \textnormal{ln} N,
\label{eq:bic}
\end{equation}
where, $\textnormal{ln} f(q,r|\theta_{i})$ denotes the log-likelihood of data under model~\eqref{eq:P_q_r_MOG} with parameter $\theta$, $n$ is the number of independent parameters to be estimated in the model and $N$ is the number of observations in the training data set.
In general, the BIC measure penalizes more complex models and favors the model which is rendered most plausible by the data at hand.
We chose the model parametrization $\theta^{*}$ that has largest BIC value in the $\theta$ search space.

\subsection{Image Quality Assessment (IQA)}
\label{IQA}
In this paper, we consider the following two image quality features of a facial image: pose and illumination.
Pose and illumination have proven record of being a strong performance predictor features for face recognition systems~\cite{beveridge2008focus,phillips2013existence}.
Therefore, these two image quality features are sufficient to demonstrate the merit of the proposed performance prediction model.
Furthermore, the choice of these two quality parameters is also motivated by the availability of a public facial image data set (\ie MultiPIE~\cite{gross2008multipie}) containing systematic pose and illumination variations.
Based on the classification scheme for facial image quality variations proposed by the ISO~\cite{iso_iec_29794-5:2010}, head pose and illumination correspond to subject characteristics and acquisition process characteristics respectively.
Furthermore, both quality parameters correspond to dynamic characteristics of a facial image as described in~\cite{iso_iec_29794-5:2010}.

\begin{figure}
 \centering
 \includegraphics[width=\linewidth]{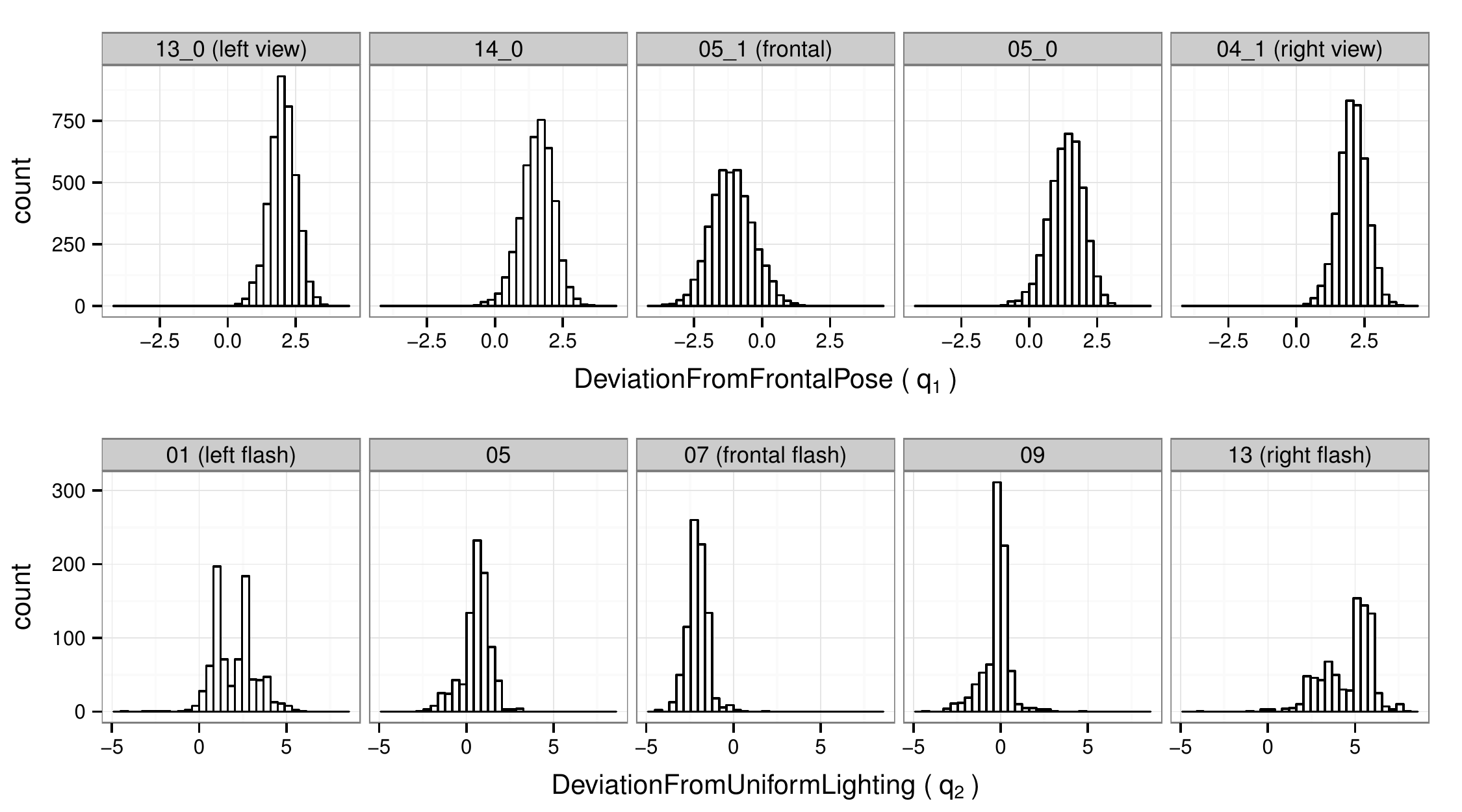}
 \caption{Distribution of image quality features measured by COTS-IQA on the MultiPIE training data set.}
 \label{fig:mpie_train_q1q2_dist_N10_m1.pdf}
\end{figure}

In this paper, we use a Commercial off-the-shelf (COTS) Image Quality Assessment (IQA) tool \texttt{dbassess} by~\cite{facevacs2010}.
Now onwards, we refer to this IQA using the acronym COTS-IQA.
The distribution of \verb+DeviationFromFrontalPose+ ($q_{1}$) and \verb+DeviationFromUniformLighting+ ($q_{2}$) quality features measured by COTS-IQA for the first fold (of 10-fold validation) training data set from MultiPIE data set is shown in~\figurename\ref{fig:mpie_train_q1q2_dist_N10_m1.pdf}.
The distribution of $q_1$ for frontal view images is centered around $-1.0$ while for non-frontal views, it shifts toward $+2.0$.
Similarly, while keeping the pose fixed to frontal view, we vary the illumination and observe that for frontal illumination the distribution of $q_2$ is centered around $-2.0$ while for other illumination conditions it shifts towards values $\geq 0$.
These distributions show that although COTS-IQA is accurate, its quality feature measurements are biased -- both left and right profile views are mapped to similar range of values thereby loosing the distinction between the two types of profile views.

\begin{figure}
 \centering
 \includegraphics[width=0.8\linewidth]{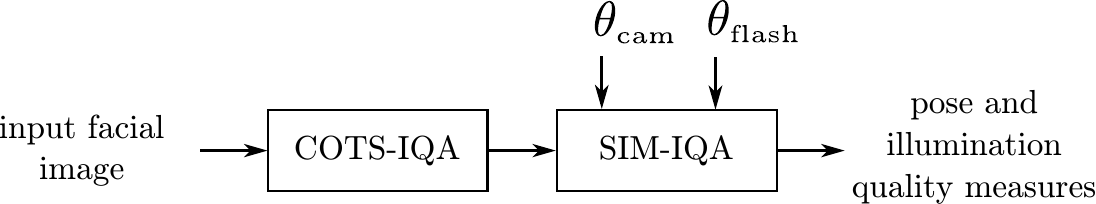}
 \caption{Input/Output features of the unbiased IQA (SIM-IQA) derived from COTS-IQA.}
 \label{fig:sim_iqa_block_dia.pdf}
\end{figure}

\begin{figure}[h]
 \centering
 \includegraphics[width=\linewidth]{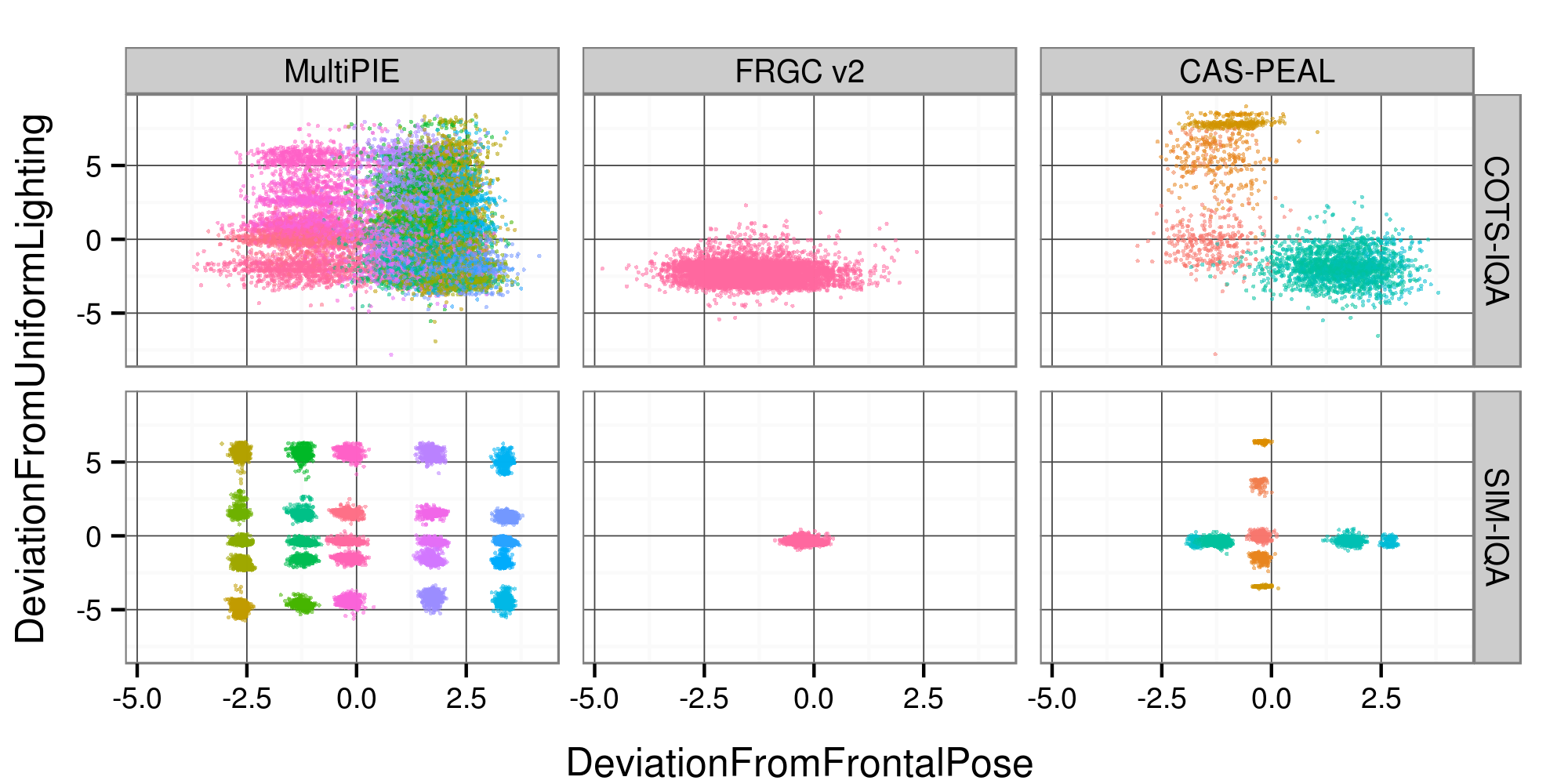}
 \caption{Quality space of COTS-IQA and unbiased IQA (SIM-IQA) which is derived from the COTS-IQA.}
 \label{fig:COTS-IQA_SIM-IQA_illustration.jpg}
\end{figure}

To demonstrate the performance prediction capability achievable by an accurate and unbiased IQA, we derive an unbiased IQA from the COTS-IQA -- henceforward referred using the acronym SIM-IQA.
The SIM-IQA is derived from the COTS-IQA and uses ground truth camera and flash positions, as shown in~\figurename\ref{fig:sim_iqa_block_dia.pdf}, to achieve more accuracy and unbiased quality assessments as shown in~\figurename\ref{fig:COTS-IQA_SIM-IQA_illustration.jpg}.
While the ground truth camera and flash positions are sufficient to simulate an IQA, we use input from COTS-IQA in order capture the characteristics of a realistic IQA.
It is important to understand that we do not use any other specific properties of COTS-IQA tool and therefore any other IQA tool can be easily plugged into this model.

SIM-IQA is obtained by transforming quality feature measurement by COTS-IQA such that images captured by same camera and under same flash map to distinctly separated clusters in the quality space as shown in~\figurename\ref{fig:COTS-IQA_SIM-IQA_illustration.jpg}. Let $\mathbf{A}=\{ \mathbf{a}_{i} \}$, where $\mathbf{a}_{i}=[q_{1}, q_{2}, \gamma_{1}, \gamma_{2}] \in \mathbb{R}^{4}$ such that $q_{1}, q_{2}$ are the pose and illumination quality measurements by COTS-IQA and $\gamma_{1}, \gamma_{2}$ be the corresponding ground truth camera and flash angle with frontal view as the reference (supplied with the image data set) for the $i^{\textnormal{th}}$ facial image sample. The corresponding quality measurements by SIM-IQA is $\mathbf{B}=\{ \mathbf{b}_{i} \}$, where $\mathbf{b}_{i} = [\hat{q}_{1}, \hat{q}_{2}] \in \mathbb{R}^{2}$ such that
\begin{align}
\hat{q}_{1} &= a \gamma_{1} + (q_{1} - \mu_{1}^{\gamma_{1},\gamma_{2}}) \nonumber \\
\hat{q}_{2} &= b \gamma_{2} + (q_{2} - \mu_{2}^{\gamma_{1},\gamma_{2}}) \nonumber
\end{align}
where, $a=1/10, b=1/18$ are scaling factor for the angle measurements $\gamma_{\{1,2\}}$ measured in degree and $\mu_{\{1,2\}}^{\gamma_{1},\gamma_{2}}$ are the mean of $q_{\{1,2\}}$ for each unique combination of ground truth camera and flash azimuth $\gamma_{\{1,2\}}$.
From the MultiPIE training data, we create the matrices $\mathbf{A}$ and $\mathbf{B}$ and compute a transformation matrix $\mathbf{x} \in \mathbb{R}^{4 \times 2}$ such that
\begin{equation}
\argmin_{\mathbf{x}} || \mathbf{A} \mathbf{x} - \mathbf{B} || \nonumber
\end{equation}
whose optimal solution, in the least square sense, is $\mathbf{x} = (\mathbf{A}^{T}\mathbf{A})^{-1} \mathbf{A}^{T} \mathbf{B}$.

\section{Experiments}
\label{dutta2015predicting_exp}
This section deals with the experiments designed to train and test the performance prediction model described in Section~\ref{dutta2015predicting_model_description}.
The description of the facial image data sets and face recognition systems used in these experiments are presented in Section~\ref{exp:datasets} and~\ref{exp:face_recognition_systems} respectively.
Several practical aspects of training a performance prediction model are dealt in Section~\ref{ssec:exp:model_training}.
In Section~\ref{ssec:exp:perf_pred}, we evaluate the accuracy of performance predictions on a test data set that is disjoint from the training data set.

\subsection{Data sets}
\label{exp:datasets}
We use the following three publicly available facial image data sets for all our experiments: MultiPIE~\cite{gross2008multipie}, FRGC v2~\cite{phillips2005overview} and CAS-PEAL~r1~\cite{gao2008caspeal}.
\begin{figure}[h]
\begin{center}
   \includegraphics[width=0.8\linewidth]{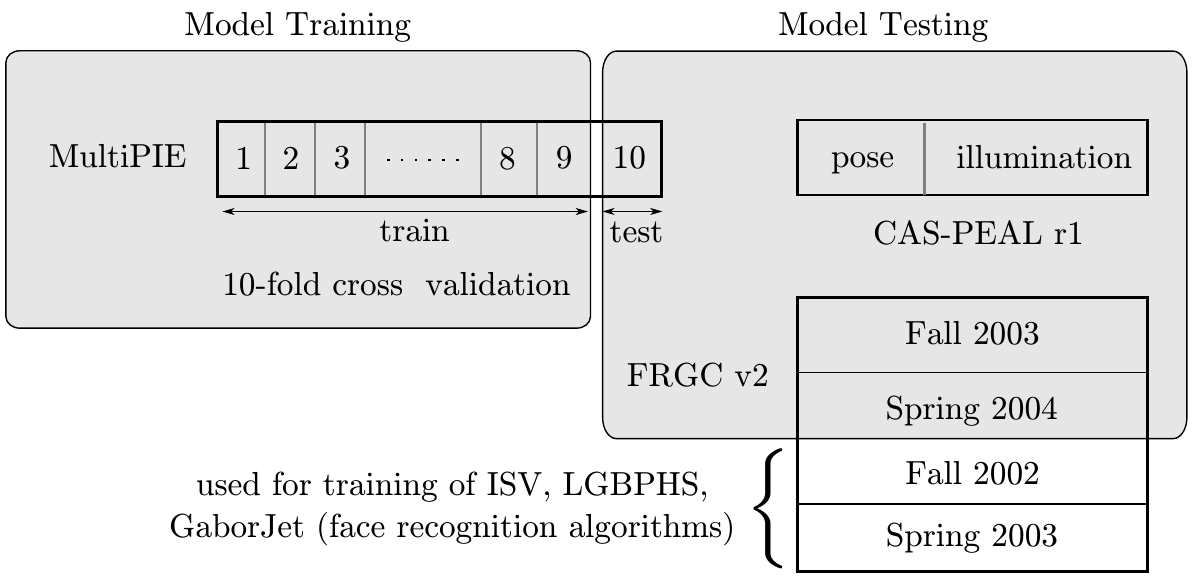}
\end{center}
   \caption{Data sets used for training and testing of the proposed model.}
\label{fig:train_test_dataset_illus.pdf}
\end{figure}

\begin{figure}[h]
 \centering
 \includegraphics[width=0.8\linewidth]{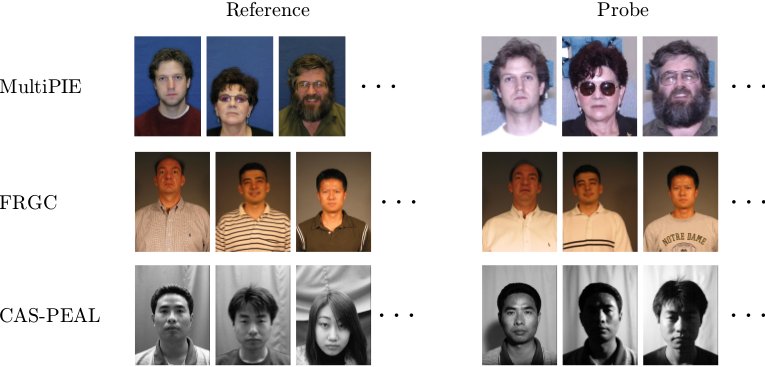}
 \caption{Sample reference and probe images from facial image data sets used in this paper.}
 \label{fig:prb_ref_img_illus.pdf}
\end{figure}

\begin{figure}
\begin{center}
   \includegraphics[width=0.8\linewidth]{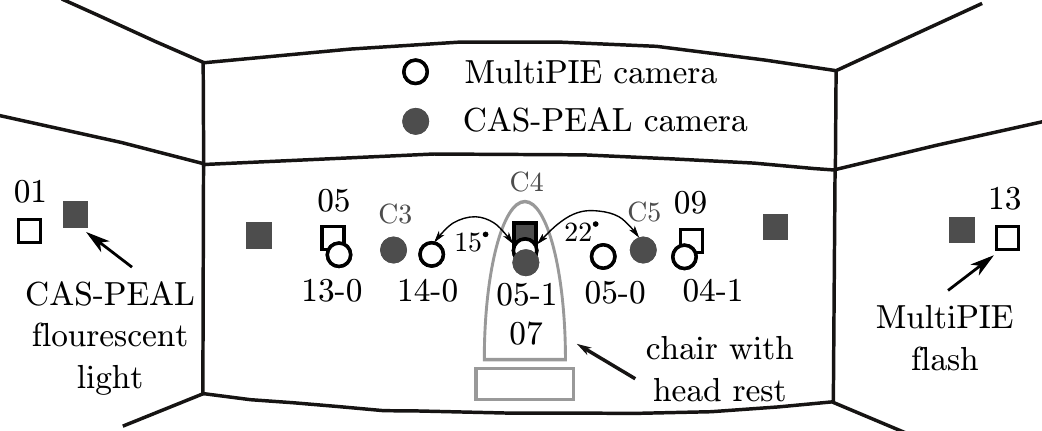}
\end{center}
   \caption{Camera and flash positions of the MultiPIE and CAS-PEAL.}
\label{fig:caspeal_mpie_capture_setup.pdf}
\end{figure}

The MultiPIE data set is primarily used for 10-fold cross validation of the model.
This data set contains systematic variations in pose and illumination and therefore has sufficient data to train and test the model based on two image quality parameters -- pose and illumination.
We use images of all the $337$ subjects across four sessions (first recording only) from the neutral expression subset of the MultiPIE.
The impact of session variation is eliminated by choosing both probe (or, query) and reference (or, enrollment) images from the same session.
The reference set contains high quality frontal mugshots and image quality variations exists only in the probe set as shown in~\figurename\ref{fig:prb_ref_img_illus.pdf}.
This simulates a real-world face verification scenario where the gallery is fixed to contain a set of high quality frontal mugshot of known individuals and facial image quality variations exists mainly in the probe set.
Recall that the proposed model can accommodate quality variation in both probe and gallery images.
However, to simulate real-world face verification scenario, we only vary the quality of probe image while keeping the gallery quality fixed.
The probe set contains $22960$ unique images of $337$ subjects captured by five camera (each separated by $15^{\circ}$) and 5 flash positions as depicted in~\figurename\ref{fig:caspeal_mpie_capture_setup.pdf}.
For the N-fold cross validation, we partition the full probe set into $N=10$ blocks such that each block contains $2296$ images randomly sampled from the full probe set.
Of the $10$ blocks, one block is retained as the validation data for testing the model while the remaining $9$ blocks are used for training the model as depicted in~\figurename\ref{fig:train_test_dataset_illus.pdf}.
This cross-validation process is repeated $10$ times such that each block is used as a test set exactly one time.
This ensures that training set has sufficient number of samples distributed in the quality space.
For each fold, the training set contains $20664$ match and $4764188$ non-match scores corresponding to $20664$ unique probe images and the testing set contains $2296$ match and $528381$ non-match scores corresponding to $2296$ unique probe images.

We also test the trained model on two other data sets that are independent from the training data set.
The first data set is the Fall 2003 and Spring 2004 subset (neutral expression, controlled condition only) of the FRGC v2 data set as shown in~\figurename\ref{fig:prb_ref_img_illus.pdf}.
This subset contains frontal view neutral expression images captured under controlled condition and therefore allows us to assess the performance prediction capability of the model on good quality images.
A single image of each subject is used as the reference while the remaining images are used as the probe.
The selected FRGC subset contains $7299$ match and $2596256$ non-match scores corresponding to $7299$ unique probe images.
Again, to minimize the impact of session variation, we chose probe and gallery images from the same session.

The second data set is the CAS-PEAL data set.
Since the model is trained to consider only pose and illumination image quality features, we include facial images of $1040$ subjects from the pose and illumination subset of the CAS-PEAL data set.
For the pose variation, we only include camera $\{C3,C5\}$ (when looking into frontal camera $C4$) corresponding to $PM\{-22^{\circ}, +22^{\circ}\}$ deviation from the frontal pose since the model is trained only on pose variations of $\pm 30^{\circ}$.
Contrary to documentation, the CAS-PEAL data set includes a pose variation of $\pm 15^{\circ}$ for some subjects.
We pool these images in the $\pm 22^{\circ}$ category.
The illumination variation subset contains images illuminated by a fluorescent (F) and incandescent (L) light source with elevation of $0^{\circ}$ and the following variations in azimuth: $IFM\{-90^{\circ}, -45^{\circ}, 0^{\circ}, 45^{\circ}, 90^{\circ} \}$.
Recall that the training data (based on MultiPIE) contained camera flash as the illumination source.
These camera and illumination positions of the CAS-PEAL data set are also depicted in~\figurename\ref{fig:caspeal_mpie_capture_setup.pdf}.

\subsection{Face Recognition Systems}
\label{exp:face_recognition_systems}
The impact of image quality variations on recognition performance also varies according to the capabilities of the face recognition system under consideration.
Therefore, we train and test the proposed model on the following six face recognition systems that have varying levels of tolerance towards facial image quality variations: FaceVACS~\cite{facevacs2010}, Verilook~\cite{verilook2011}, Cohort LDA~\cite{bolme2012csu}, Inter-Session Variability modeling~\cite{wallace2011intersession}, Gabor-Jet~\cite{guenther2012disparity}, Local Gabor Binary Pattern Histogram Sequences~(LGBPHS)~\cite{zhang2005local}.
The first two systems are Commercial off-the-shelf (COTS) and the remaining four systems are open source face recognition systems.
Throughout this paper, we refer to these six face recognition systems by the abbreviations COTS-A, COTS-B, cLDA, ISV, GaborJet, LGBPHS respectively.
We use the implementation of ISV, GaborJet and LGBPHS available in \texttt{FaceRecLib}~\cite{gunther2012open}.
The COTS-A, COTS-B and cLDA systems are pre-trained and ISV, GaborJet, LGBPHS are trained using the Fall 2002 and Spring 2003 subset of the FRGC v2 data set as defined in the training protocol of \texttt{FaceRecLib}.
We supply the same manually annotated eye coordinates to all the six face recognition systems to ensure consistency in facial image registration even for non-frontal images.

For face recognition systems deployed in real-world, the vendor (or, the user) sets an operating point by judiciously choosing the value of the decision threshold as shown in~\figurename\ref{fig:black_box_qr_model_illus}.
This decision threshold is chosen based on the user requirement of a certain minimum False Match Rate (FMR) or False Non-Match Rate (FNMR).
For the six systems considered in this paper, we simulate such a real world setup by setting the operating point to achieve a FMR of $0.1\%$ for the first three systems and $1\%$ for the remaining three systems.
To generate the Receiver Operating Characteristics (ROC) curve of~\figurename\ref{roc_mpie_Train9_Test1_Nqs12_fv.pdf},~\ref{fig:roc_frgc_Train9mpie_Nqs12.pdf}, and ~\ref{fig:roc_caspeal_Train9mpie_Nqs12.pdf}, we train eight separate models to predict performance for the face recognition systems operating at the following FMR: $\{0.01\%,0.03\%,0.1\%,0.3\%,1\%,3\%,10\%,30\%\}$.
The corresponding decision threshold for COTS-A and COTS-B directly comes from their SDK (\ie vendor).
For the remaining four open source systems, we use frontal view and illumination $(05\_1,07)$ images of the MultiPIE training data set to compute the verification decision threshold corresponding to each FMR.

\subsection{Model Training}
\label{ssec:exp:model_training}
\begin{figure*}
\centering
\subfloat[COTS-IQA]{\includegraphics[width=0.48\linewidth]{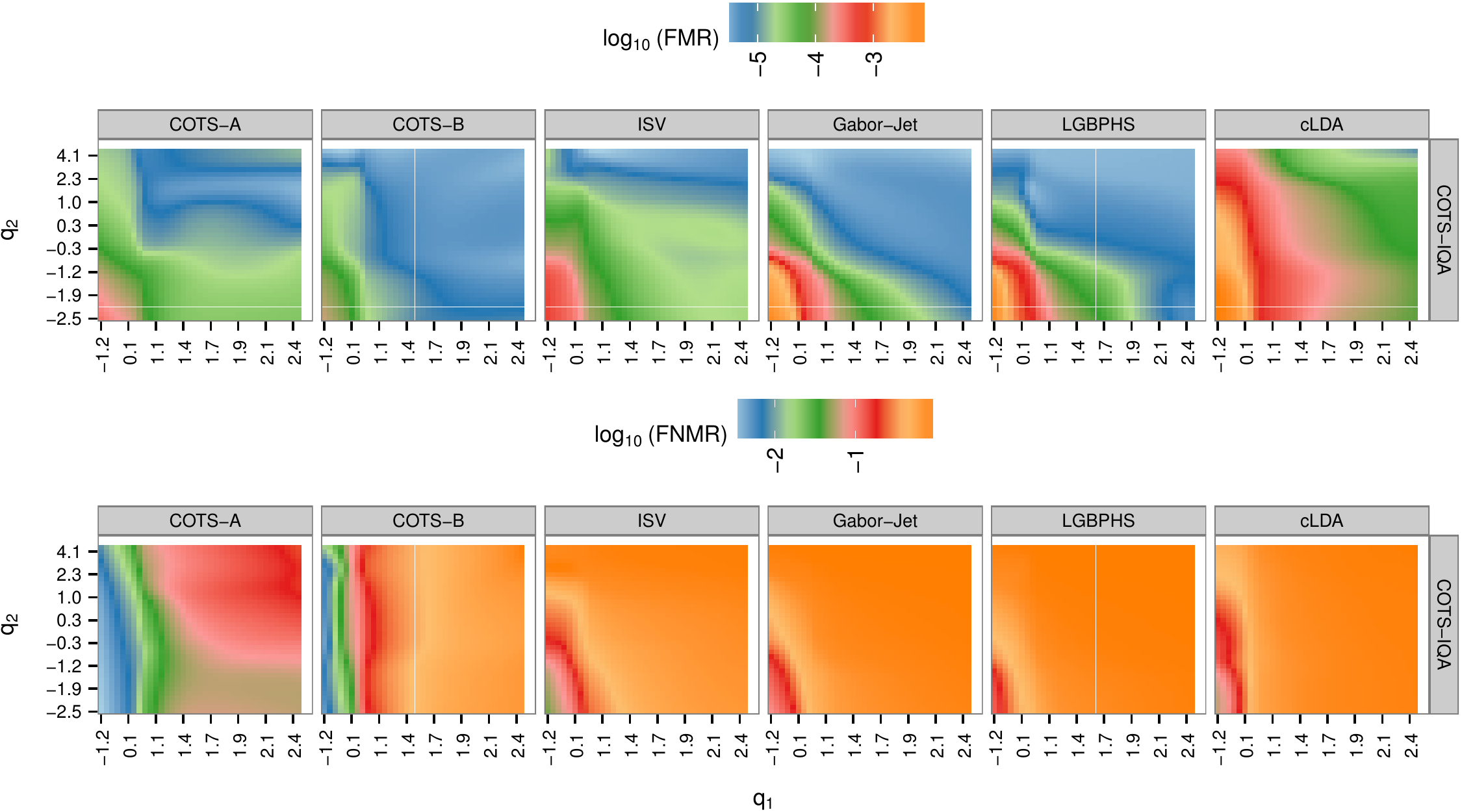}%
\label{fig:qr_map_Train9_Test1_Nqs5_K25_SigmaVVI_Nrand20_fv_IQA.pdf}} $\quad$
\subfloat[SIM-IQA]{\includegraphics[width=0.48\linewidth]{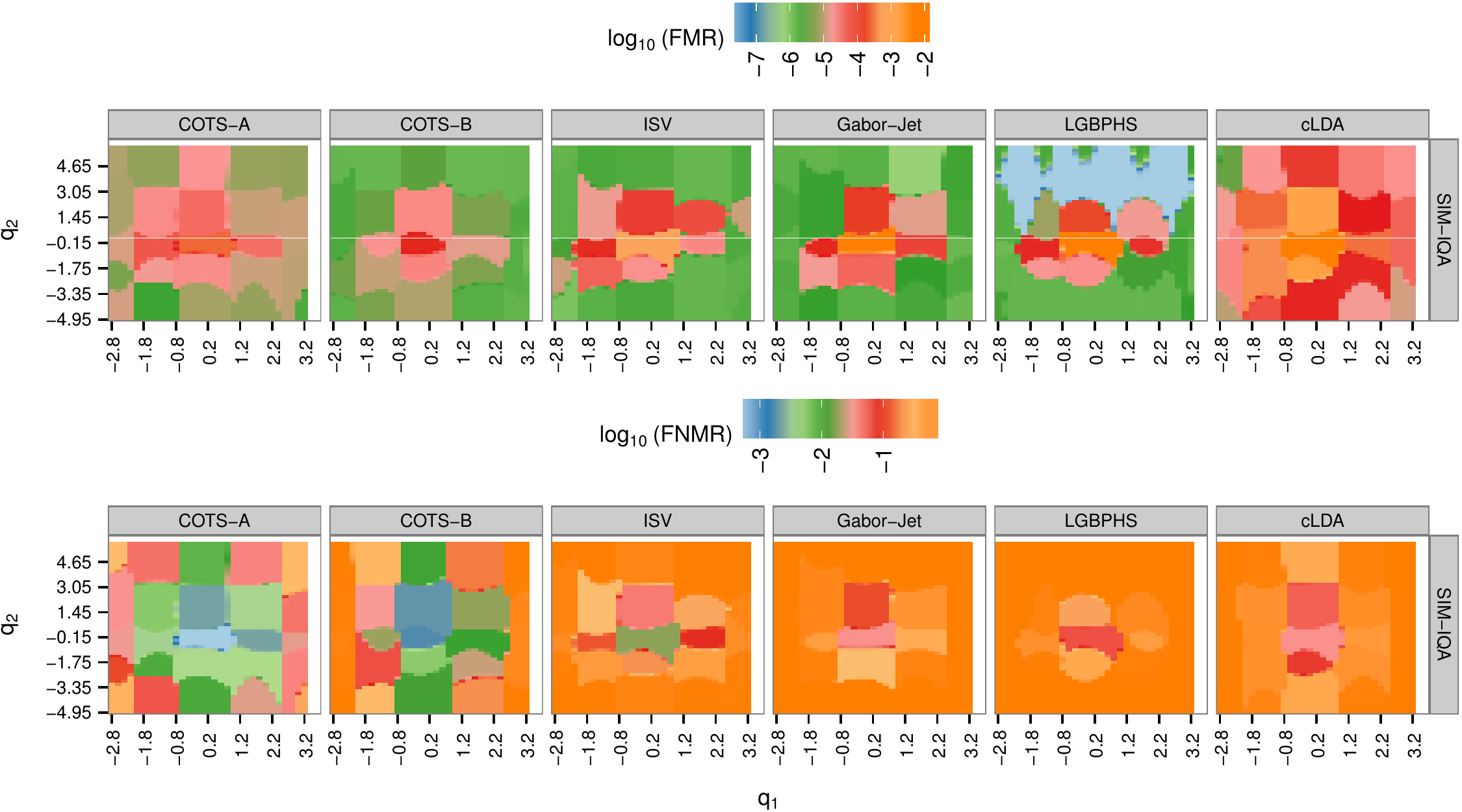}%
\label{fig:qr_map_Train9_Test1_Nqs5_K25_SigmaVVI_Nrand20_fv_IQA0.pdf}}
\caption{Visualization of the quality ($\mathbf{q}$) performance ($\mathbf{r}$) space of six face recognition systems based on two Image Quality Assessors: COTS-IQA and SIM-IQA.}
\label{fig:qr_map_Train9_Test1_Nqs5_K25_SigmaVVI_Nrand20.pdf}
\end{figure*}

We begin with a coarse sampling of the COTS-IQA quality space based on $N_{qs}=12$ quantiles of evenly spaced probabilities along each dimension of the 2D quality space.
Discarding the first and last sampling points (corresponding to quantiles with probabilities $0.0$ and $1.0$ respectively), we have $10 \times 10$ unique sampling points in the 2D quality space resulting in $100$ overlapping regions around each sampling point.
As described in Section~\ref{model_training}, we draw $N_{\textnormal{rand}}=20$ random samples of $\mathbf{q}$ and $\mathbf{r}$ in each region which results in a training data set $\tilde{\mathcal{D}}_{\textnormal{train}} \in \mathbb{R}^{2000 \times 4}$.
A small value of $N_{\textnormal{rand}}$ ensures that the training process completes quickly ($\sim 5$sec).

\begin{figure}
\begin{center} \includegraphics[width=0.7\linewidth]{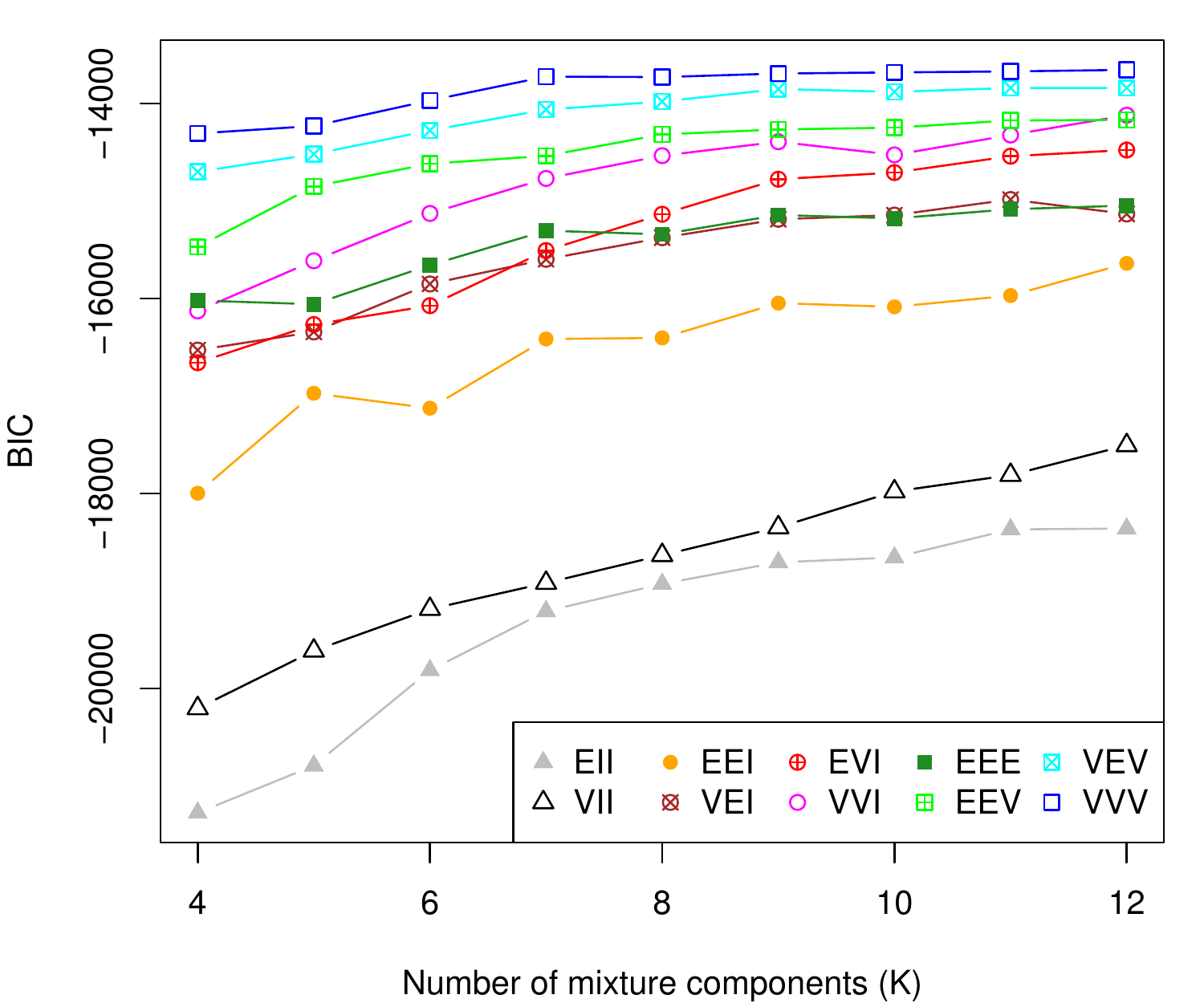}
\end{center}
   \caption{BIC value corresponding to different assignments of model parameter $\theta$.}
\label{fig:fv_BIC_vs_theta_Train9_Test1_Nbeta20_Nqs12_tid3.pdf}
\end{figure}

We select the optimal model parameters based on the BIC criterion as described in Section~\ref{model_param_selection}.
For COTS-A,~\figurename\ref{fig:fv_BIC_vs_theta_Train9_Test1_Nbeta20_Nqs12_tid3.pdf} shows the BIC value for the $\theta$ parameter search space.
We select $\theta^{*}=(9,\textnormal{VVV})$ because the BIC value attains maximum value at this point and saturates beyond it.
Furthermore, the remaining five face recognition systems also have similar trend of BIC values and therefore $\theta^{*}=(9,\textnormal{VVV})$ is selected as the model parameter for all six face recognition systems.
Here, VVV corresponds to a covariance matrix parametrization scheme in which all mixture components have varying (V) volume, the shapes and orientation of mixture components may vary (V).

The quality space sampling for SIM-IQA (\ie the unbiased IQA derived from COTS-IQA) is much simpler.
Quality regions correspond to the $25$ clusters formed by unique combination of $5$ camera and $5$ flash positions as shown in~\figurename\ref{fig:caspeal_mpie_capture_setup.pdf}.
With $N_{\textnormal{rand}}=20$, the training set corresponding to SIM-IQA is $\tilde{\mathcal{D}}_{\textnormal{train}} \in \mathbb{R}^{500 \times 4}$.
To model this nicely clustered quality space, the optimal model parametrization is manually selected to $\theta^{*} = (25,\textnormal{VVI})$.

Using the Expectation Maximization (EM) algorithm implementation available in the \texttt{mclust}~\cite{fraley2012mclust} library, we learn the model parameters of the GMM~(\ref{eq:P_q_r_MOG}) for both COTS-IQA and SIM-IQA. 
The QR space based on COTS-IQA and SIM-IQA as learned by the model is shown in~\figurename\ref{fig:qr_map_Train9_Test1_Nqs5_K25_SigmaVVI_Nrand20.pdf}.
In this figure, the X and Y axis correspond to \verb+DeviationFromFrontalPose+ ($q_{1}$) and \verb+DeviationFromUniformLighting+ ($q_{2}$) respectively of the probe image and the color denotes the value of FMR and FNMR at the operating point obtained using~(\ref{eq:expectation_r_given_q0}).
The QR space based on SIM-IQA as shown in~\figurename\ref{fig:qr_map_Train9_Test1_Nqs5_K25_SigmaVVI_Nrand20_fv_IQA0.pdf} contains small patches which correspond to distinctly separated quality clusters present in the quality space of SIM-IQA as shown in~\figurename\ref{fig:COTS-IQA_SIM-IQA_illustration.jpg}.
The conditional expectation values are shown in log$_{10}$ scale.

\figurename\ref{fig:qr_map_Train9_Test1_Nqs5_K25_SigmaVVI_Nrand20.pdf} is quite revealing in several ways.
First, it shows the complex nature of the landscape formed by face recognition performance measures like FNMR and FMR even in a simple two dimensional quality space.
From this map, it is evident that image quality variations (\ie pose and illumination) in the probe image have strong impact on the performance of all the six face recognition systems.
Second, all the six face recognition systems show asymmetric recognition performance variations along the pose $q_{1}$ and illumination $q_{2}$ quality axes.
For instance,~\figurename\ref{fig:qr_map_Train9_Test1_Nqs5_K25_SigmaVVI_Nrand20.pdf} clearly shows that performance on right side facial view $(q_{1} > 0)$ is distinctly higher than the left side view $(q_{1} < 0)$.
Furthermore, such asymmetric performance variations also exist for illumination direction corresponding to left and right side view of face.

\subsection{Performance Prediction}
\label{ssec:exp:perf_pred}
The test data set in each fold of the 10-fold cross validation set contains the following record for each verification attempt: similarity score $s$, quality of probe image $\mathbf{q}$, and ground truth for verification decision (match or non-match).
The trained model can predict recognition performance $\mathbf{r}$ based solely on the quality of the probe images $\mathbf{q}$.
However, the test data set does not contain the value of true recognition performance measure per verification attempt.
Therefore, we resort to assessing the merit of model based performance predictions by pooling results according to the ground truth camera and flash label of each probe image.
For each ground truth quality pool (\ie camera-id and flash-id), we compute the nature of FMR and FNMR distribution using the Beta distribution model of Section~\ref{prob_model_q_r}.
The mean and $95\%$ credible interval defines the variation in true recognition performance over the ground truth quality pool.
We also accumulate all model predictions of the recognition performance (FMR and FNMR) corresponding to each ground truth quality pool.
The mean of these predictions and $95\%$ confidence interval (difference between $0.975$ and $0.025$ quantiles) define the variation in predicted performance over each ground truth quality pool.
We train the model at several operating points and plot the full ROC curves corresponding to the true and model predicted (both using COTS-IQA and SIM-IQA) performance.
Due to space contraints, we only show the true and predicted ROC curves corresponding to the MultiPIE test set for the COTS-A system in~\figurename\ref{roc_mpie_Train9_Test1_Nqs12_fv.pdf}.
The remaining five face recognition systems follow similar trends.
Such plots for the remaining five face recognition systems are included in the supplementary material accompanying this paper.
The ROC curves are pooled according to the ground truth camera-id and flash-id of the probe images.
We only show the $95\%$ confidence and credible interval for FNMR to aid proper visualization of the results.

We also evaluate the merit of predicted performance on a test set derived from the FRGC data set and the CAS-PEAL pose and illumination subset.
For these evaluations on independent data set, the model was trained on the MultiPIE training set corresponding to the first fold of 10-fold cross validation. 
The FRGC test set contains only frontal view images and therefore~\figurename\ref{fig:roc_frgc_Train9mpie_Nqs12.pdf} shows a single pool of quality.
For the CAS-PEAL test set, the true and predicted recognition performance for pose variations are shown in~\figurename\ref{fig:roc_caspeal_Train9mpie_Nqs12.pdf}.
Similar plot for illumination variation is present in the supplementary material accompanying this paper.

\subsubsection*{Error versus Reject Curve (ERC)} 
The authors of~\cite{grother2007performance} have proposed the Error versus Reject Curve (ERC) as a metric for evaluating the efficacy of a performance prediction system in identifying samples that contribute to verification error (\ie FNMR or FMR).
The ERC plots FNMR (or FMR) against the fraction of verification attempts rejected by a performance prediction system as being ``poor'' quality.
A biometric system with FNMR of $x$ would achieve a FNMR of $0$ by rejecting all the $x$ verification attempts that would lead to a False Non-Match.
This provides a benchmark for evaluating performance prediction systems.

\begin{figure}[t]
 \centering \includegraphics[width=\linewidth]{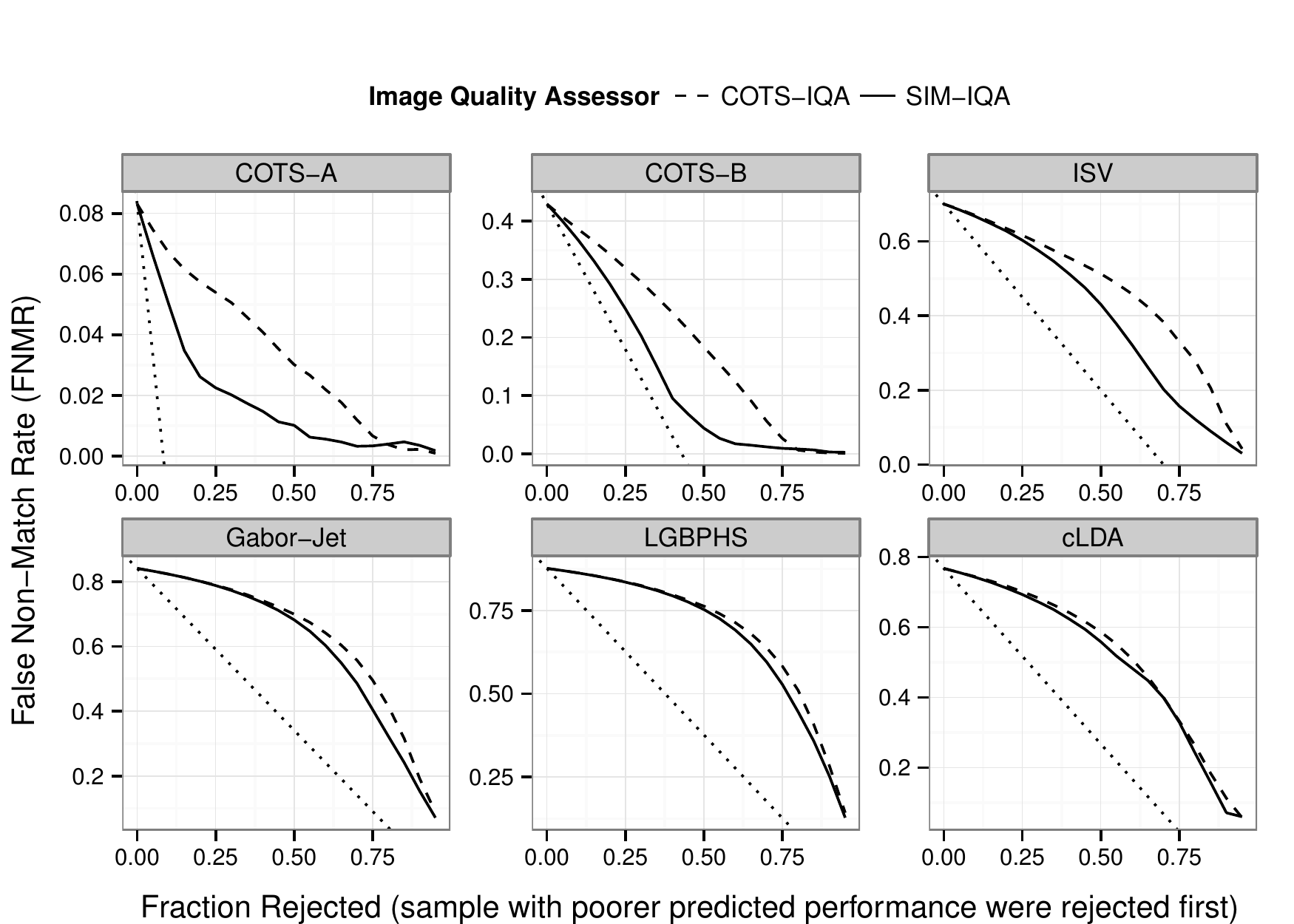}
 \caption{Error versus reject curve for the proposed performance prediction model based on two different Image Quality Assessors (IQA). Note that the fine dotted line denotes a sample rejection scheme based on an ideal performance prediction system (the benchmark).}
 \label{fig:erc_fnmr_mpie_Train9_Test1_K9_SigmaVVV_Nbeta20_Nqs12_tid333555.pdf}
\end{figure}

In this paper, we also use the ERC to evaluate the merit of performance predictions made by our model.
We sort all the verification attempts accumulated from the test set of N-fold cross validation based on the corresponding FNMR predicted by our model.
We sequentially remove verification attempts -- verification attempts with poorest predicted performance are rejected first -- and recompute the FNMR to create the ERC plot shown in~\figurename\ref{fig:erc_fnmr_mpie_Train9_Test1_K9_SigmaVVV_Nbeta20_Nqs12_tid333555.pdf}.

As expected, the ERC plot of~\figurename\ref{fig:erc_fnmr_mpie_Train9_Test1_K9_SigmaVVV_Nbeta20_Nqs12_tid333555.pdf} shows that performance predictions made using SIM-IQA are more accurate.
Furthermore, for COTS-A and COTS-B, we observe an initially sharp decline in FNMR which indicates that the model is good at identifying the poorest quality samples in pose and illumination quality space.
The flattening out of the ERC curves after the initial sharp decline suggests that pose and illumination quality features are not sufficient to identify ``poor'' quality samples containing image quality degradations along other quality dimensions.
We require additional image quality features to capture all the quality variations present in the test set.
For the remaining four face recognition systems, the ERC curves remain flattened until a majority of samples are rejected.
The reason for this is explained by the composition of our test data set and the nature of these systems.
The test set used for generating these ERC curve contains almost $80\%$ non-frontal images (only $4606$ of $22964$ probe images are frontal).
The four face recognition systems are known to be highly sensitive to even small pose variations, and therefore a large number of non-frontal samples have to be rejected to bring down the FNMR.
On the contrary, COTS-A and COTS-B have some tolerance towards small deviation from frontal pose (like camera $14\_0$ and $05\_0$) and therefore significant drop in FNMR is achieved after rejecting a small number of extreme non-frontal images (corresponding to camera $13\_0$ and $04\_1$).

\section{Discussion}
\label{dutta2015predicting_discussion}
\begin{figure}[t]
 \centering
 \includegraphics[width=\linewidth]{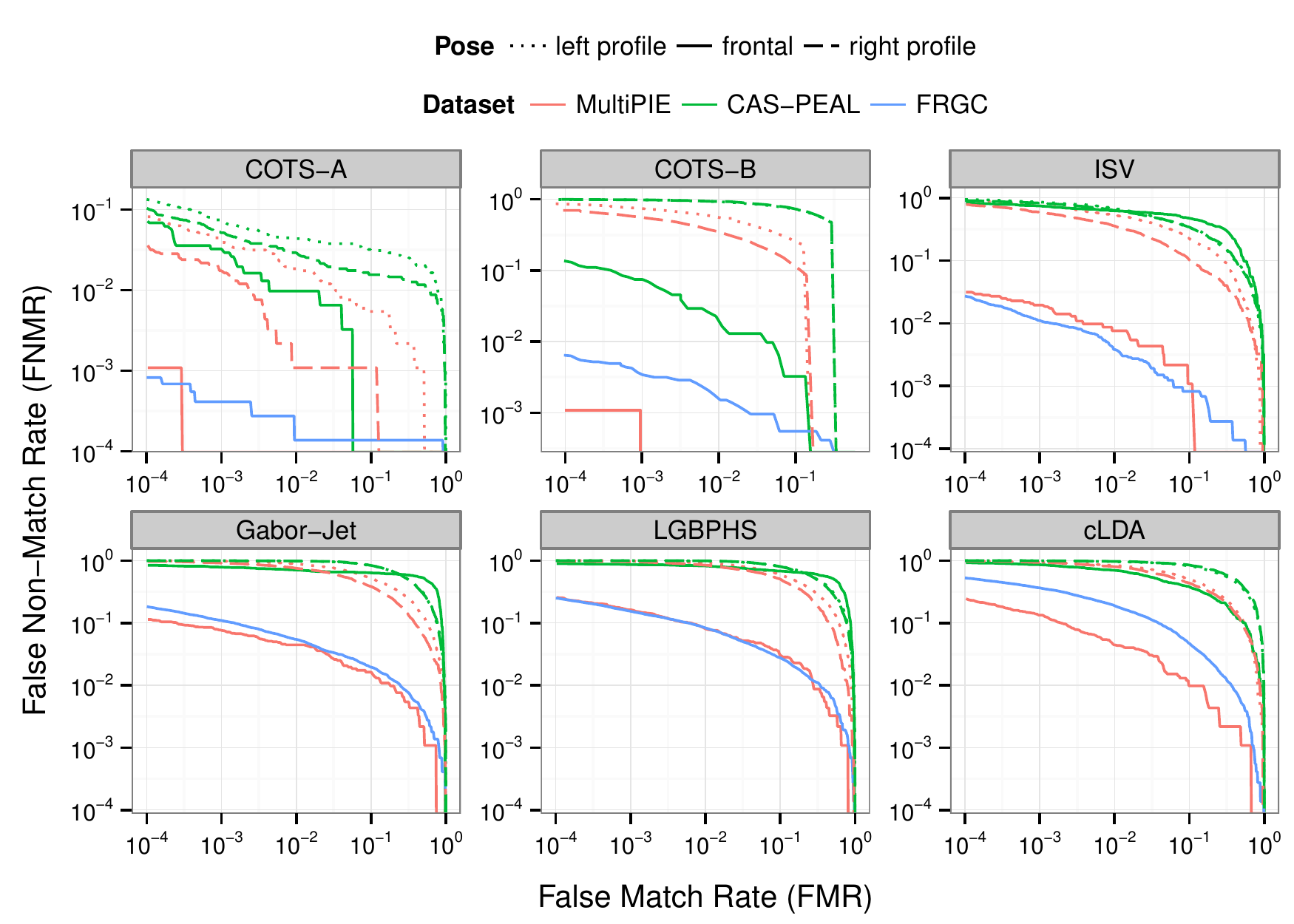}
 \caption{The nature of face recognition systems towards difference in facial pose (left and right side views) and the differences across independent facial image data sets. Note: left and right view correspond to MultiPIE camera $\{13\_0,04\_1\}$ \ie $(-30^{\circ}, +30^{\circ})$ and CAS-PEAL camera $\{C2,C6\}$ \ie $(-45^{\circ}, +45^{\circ})$.}
 \label{fig:frs_roc_pose_dataset.pdf}
\end{figure}

\begin{figure*}
 \centering
 \includegraphics[width=0.7\linewidth]{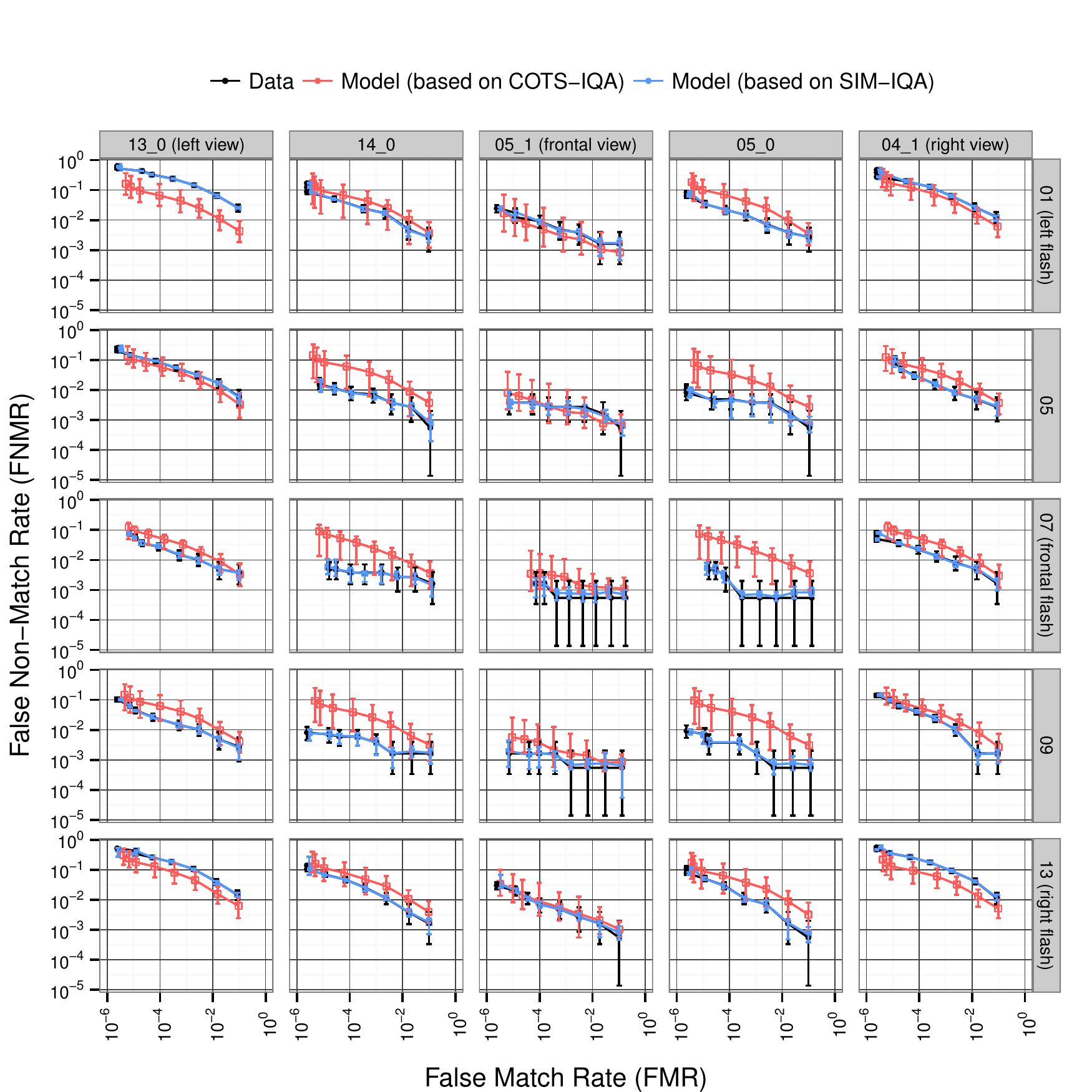}
 \caption{Face recognition performance prediction for COTS-A system using model based on COTS-IQA and SIM-IQA for MultiPIE test set pooled from 10-fold cross validation.}
 \label{roc_mpie_Train9_Test1_Nqs12_fv.pdf}
\end{figure*}

\figurename\ref{fig:frs_roc_pose_dataset.pdf} is central to the discussions presented in this section.
Therefore, we first describe the contents of this figure.
We compare the performance of face recognition systems when operating on left and right side view images corresponding to probe image subsets taken from the MultiPIE and CAS-PEAL data set.
We observe that all the face recognition systems have better recognition performance while comparing probe images containing the right side facial view.
This recognition performance bias is most distinctly visible in \figurename\ref{fig:frs_roc_pose_dataset.pdf} for COTS-A system operating on the MultiPIE data subset.
To capture such asymmetric relationship between facial pose and recognition performance, we require an IQA tool that maps left and right profile views to distinctly different regions of the quality space.
\figurename\ref{fig:mpie_train_q1q2_dist_N10_m1.pdf} shows that COTS-IQA maps both left and right profile to the same region ($q[1] \sim 2$) of the quality space thereby introducing ambiguity between left and right side view in the quality space.
On the other hand, the SIM-IQA is designed to map left and right side views to distinctly well separated regions of quality space as shown in~\figurename\ref{fig:COTS-IQA_SIM-IQA_illustration.jpg}.
Therefore, we expect the performance predictions based on COTS-IQA to have larger errors to non-frontal views and those based on SIM-IQA to be more accurate.
In \figurename\ref{roc_mpie_Train9_Test1_Nqs12_fv.pdf}, we show the true and model predicted (based on COTS-IQA and SIM-IQA quality assessors) recognition performance of COTS-A face recognition system.
As expected, we observe that model predictions based on COTS-IQA are further away from the true performance for non-frontal views while predictions based on SIM-IQA are more accurate even for non-frontal views.
The accuracy of performance predictions of the remaining five face recognition systems follow a similar trend and are shown in the supplementary material accompanying this paper.
This shows that an unbiased IQA is essential for accurate performance predictions.

\begin{figure*}
 \centering
 \includegraphics[width=0.7\linewidth]{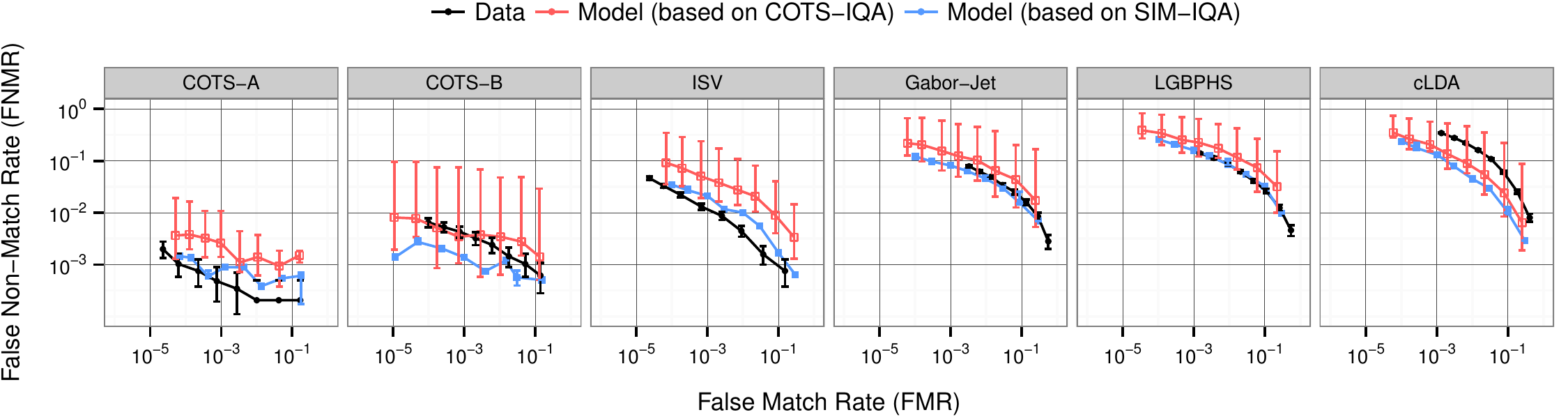}
 \caption{Model predicted and true recognition performance for test set based on the FRGC v2 data set.}
 \label{fig:roc_frgc_Train9mpie_Nqs12.pdf}
\end{figure*}

\begin{figure*}
 \centering
 \includegraphics[width=0.7\linewidth]{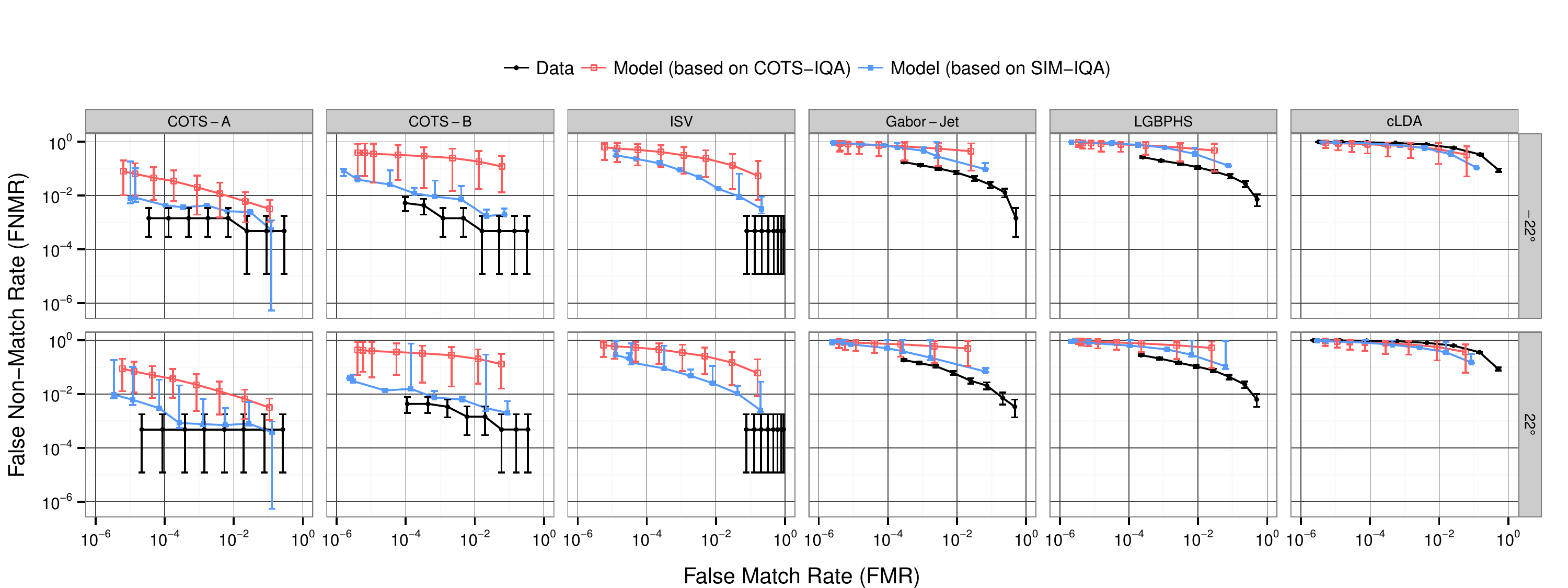}
 \caption{Model predicted and true recognition performance for pose variation test set based on the CAS-PEAL data set.}
 \label{fig:roc_caspeal_Train9mpie_Nqs12.pdf}
\end{figure*}

In our performance prediction model, we have only considered two image quality features: pose and illumination.
However, there may exist variability in the \textit{unaccounted quality space} which is formed by image quality features other than pose and illumination.
For example, in this context, the unaccounted quality space may be composed of image quality features like resolution, capture device characteristics, facial uniqueness, \etc.
Furthermore, in a controlled data set like MultiPIE, FRGC or CAS-PEAL, it is reasonable to assume that variability in unaccounted quality space within a data set remains constant while such variability differs among the data sets.
Often variability among data sets is caused by difference in the capture device and capture environment.
Such variability is the reason why some data sets are more challenging than others in the context of face recognition.
To investigate the extent of variation present in the unaccounted quality space of the MultiPIE, FRGC and CAS-PEAL data sets, we compare the recognition performance of six face recognition systems on frontal view and frontal illumination images of these data sets in~\figurename\ref{fig:frs_roc_pose_dataset.pdf}.
Since we have selected the frontal pose and illumination subset of these controlled data sets, any performance difference for a particular face recognition system can be attributed to the variability present in the unaccounted quality space of these data sets.
Furthermore, since the performance prediction model is trained solely on the MultiPIE subset, in the following analysis, we assume the unaccounted quality space of MultiPIE frontal subset to be the reference.
In~\figurename\ref{fig:frs_roc_pose_dataset.pdf}, we observe that the performance of all six face recognition systems are consistently much poorer on the CAS-PEAL data set as compared to the corresponding recognition performance on MultiPIE.
This shows that the unaccounted quality space of CAS-PEAL data set is significantly different from that of the MultiPIE or FRGC data set.
Therefore, we expect that a performance prediction model trained solely on the MultiPIE data set (using the SIM-IQA) will perform poorer on the CAS-PEAL data set.
As expected, \figurename~\ref{fig:roc_caspeal_Train9mpie_Nqs12.pdf} confirms that performance predictions (using models trained on MultiPIE subset with COTS-IQA or SIM-IQA) on the CAS-PEAL data set are erroneous because of the large difference in the unaccounted quality space variability of CAS-PEAL as compared to that of the MultiPIE data set.
Surprisingly,~\figurename\ref{fig:frs_roc_pose_dataset.pdf} reveals that there is very small difference between the performance corresponding to MultiPIE and FRGC for Gabor-Jet and LGBPHS systems while the performance is significantly different for the remaining four face recognition systems.
This suggests that while there is difference in the variability of the unaccounted quality space between FRGC and MultiPIE data sets, the Gabor-Jet and LGBPHS systems are tolerant to this difference.
Therefore, for the Gabor-Jet and LGBPHS systems, we expect a performance prediction model trained solely on the MultiPIE data set (using the SIM-IQA) will make more accurate predictions on the FRGC data set.
Furthermore, the prediction error on the FRGC data set will be high for COTS-B and cLDA because these systems are highly sensitive to the difference in unaccounted quality space of FRGC data set as shown in~\figurename\ref{fig:frs_roc_pose_dataset.pdf}.
The performance predictions on the FRGC data set are shown in Figure~\ref{fig:roc_frgc_Train9mpie_Nqs12.pdf} and the accuracy of the predictions are exactly as we expected -- more accurate predictions for Gabor-Jet and LGBPHS while high prediction error on COTS-B and cLDA.
These findings suggest that reliability of performance predictions is highly dependent on the variability that exists in the unaccounted quality space.
Therefore, to make accurate predictions for a face recognition system, we must consider all the image quality features that have an influence on the performance of that system.

The CAS-PEAL data set mainly consists of subjects from East Asia.
There is evidence that face recognition algorithms (like the six systems used in this paper) trained mainly on Caucasian faces are less accurate when applied to East Asian faces~\cite{phillips2011otherrace}.
This suggests that race of subjects contained in a verification attempt is potentially an important image quality feature (static subject characteristics according to ~\cite{iso_iec_29794-5:2010}) that is essential to address such a performance bias present in existing face recognition systems.

\section{Conclusion}
\label{dutta2015predicting_conclusion}
In this paper, we present a generative model to capture the relation between image quality features $\mathbf{q}$ (\eg pose, illumination, \etc) and face recognition performance $\mathbf{r}$ (\eg FMR and FNMR at operating point).
Such a model allows performance prediction even before the actual recognition has taken place because the model is based solely on image quality features.
A practical limitation of such a data driven generative model is the limited nature of training data.
To address this limitation, we have developed a Bayesian approach to model the nature of FNMR and FMR distribution based on the number of match and non-match scores in small regions of the quality space.
Random samples drawn from the models provide the initial data essential for training the generative model $P(\mathbf{q},\mathbf{r})$.

We evaluated the accuracy of performance predictions based on the proposed model using six face recognition systems operating on three independent data sets.
The evidence from this study suggests that the proposed performance prediction model can accurately predict face recognition performance using an accurate and unbiased Image Quality Assessor (IQA).
An unbiased IQA is essential to capture all the complex behaviours of face recognition systems.
For instance, our results show that the performance of some face recognition systems on right side view is better than the recognition performance on left side view.
Such a complex and unexpected behaviour can only be captured by an IQA that maps left and right views to different regions of the quality space.

We also investigated the reason behind high performance prediction error when the performance prediction model is applied to other independent data.
We found variability in the \textit{unaccounted quality space} -- the image quality features not considered by the IQA -- as the major factor causing inaccuracies in predicted performance.
Even controlled data sets have large amount of variability in the unaccounted quality space.
Furthermore, face recognition systems differ in their tolerance towards such variability.
Therefore, in general, to make accurate predictions on a diverse test data set, we should either consider all the relevant image quality features in order to minimize the variability in unaccounted quality space or use a classifier that is agnostic to variability in the unaccounted quality space.

This work has pointed out future work in many directions.
Clearly, the most significant effort needs to be concentrated in the direction of discovering novel features that can summarize a large number of image quality variations.
This is essential for limiting the variations present in the unaccounted quality space.
Furthermore, there is a clear need to develop accurate and unbiased Image Quality Assessment systems.
Although our model can accept image quality measurements from off-the-shelf and uncalibrated quality assessment systems, more transparent and standardized quality metrics are needed to facilitate standardized exchange of image quality information as proposed in~\cite{iso_iec_29794-1:2009}.
Future work could also investigate methods to directly incorporate probabilistic models of quality and recognition performance into the EM based training procedure.
It would also be interesting to apply the proposed model to predict the performance of other biometric systems and other classifiers in general.


\ifCLASSOPTIONcaptionsoff
  \newpage
\fi



%



\bibliographystyle{plain}
\bibliography{dutta_pami_references.bib}
\end{document}